%% file: main.tex
\documentclass{article}

\usepackage{microtype}
\usepackage{graphicx}
\usepackage{caption}
\usepackage{amsmath}
\usepackage{amssymb}
\usepackage{mathtools}
\usepackage{amsthm}
\usepackage{amsfonts}
\usepackage{nicefrac}
\usepackage{float}
\usepackage{booktabs}
\usepackage[dvipsnames]{xcolor}
\usepackage{colortbl}
\usepackage{algorithm}
\usepackage[noend]{algorithmic}
\usepackage{bbding}
\usepackage{fancyvrb}
\usepackage{listings}
\usepackage{subcaption}
\usepackage{wrapfig}

\usepackage[title]{appendix}

\DeclareCaptionLabelFormat{custom}
{%
      \textit{#1 #2.}
}

\DeclareCaptionLabelSeparator{custom}{}

\makeatletter
\def\blfootnote{\gdef\@thefnmark{}\@footnotetext}
\makeatother

\usepackage{multirow}

\usepackage[preprint]{corl_2022} 

\title{Graph Inverse Reinforcement Learning \\ from Diverse Videos
}

\author{
  Sateesh Kumar \quad Jonathan Zamora* \quad Nicklas Hansen* \\
  \quad \textbf{Rishabh Jangir} \quad \textbf{Xiaolong Wang}\\
   UC San Diego \\
}

\begin{document}
\maketitle

\begin{figure}[H]
    \centering
    \vspace{-0.3in}
    \hspace{-0.1in}
    \includegraphics[width=0.85\textwidth]{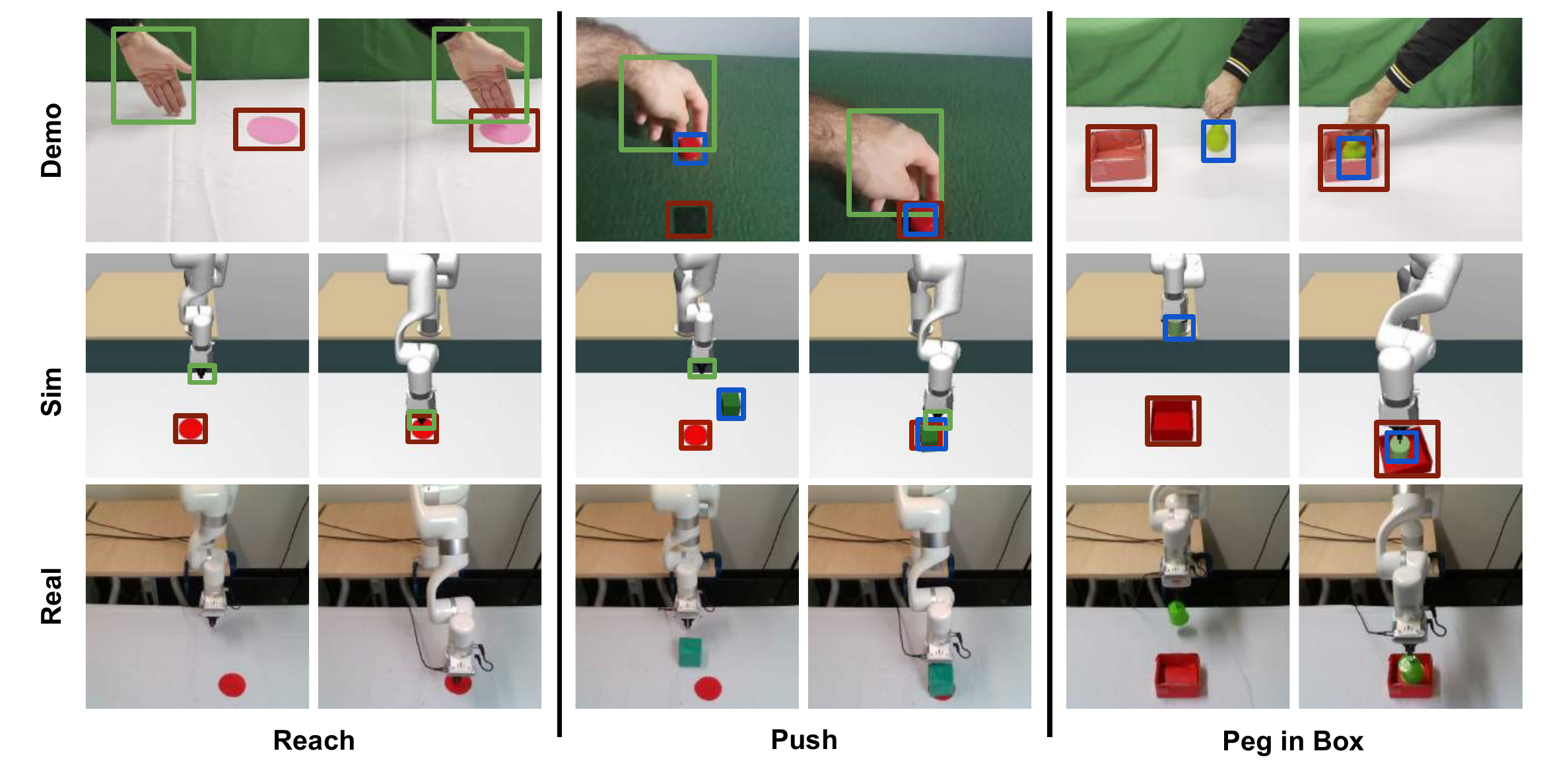}
    \caption{\textbf{GraphIRL.} We propose an approach for performing inverse reinforcement learning from diverse third-person videos via graph abstraction. Based on our learned reward functions, we successfully train image-based policies in simulation and deploy them on a real robot.}
    \label{fig:teaser}
    \vspace{-0.05in}
\end{figure}

\begin{abstract}
Research on Inverse Reinforcement Learning (IRL) from third-person videos has shown encouraging results on removing the need for manual reward design for robotic tasks. However, most prior works are still limited by training from a relatively restricted domain of videos. In this paper, we argue that the true potential of third-person IRL lies in increasing the diversity of videos for better scaling. To learn a reward function from diverse videos, we propose to perform graph abstraction on the videos followed by temporal matching in the graph space to measure the task progress. Our insight is that a task can be described by entity interactions that form a graph, and this graph abstraction can help remove irrelevant information such as textures, resulting in more robust reward functions. We evaluate our approach, \emph{GraphIRL}, on cross-embodiment learning in X-MAGICAL and learning from human demonstrations for real-robot manipulation. We show significant improvements in robustness to diverse video demonstrations over previous approaches, and even achieve better results than manual reward design on a real robot pushing task. Videos are available at \url{https://sateeshkumar21.github.io/GraphIRL}.



\end{abstract}

\section{Introduction}
\label{sec:introduction}
\blfootnote{ \vspace{-0.2in} * indicates equal contribution.}
\vspace{-0.07in}
Deep Reinforcement Learning (RL) is a powerful general-purpose framework for learning behavior policies from high-dimensional interaction data, and has led to a multitude of impressive feats in application areas such as game-playing \citep{Mnih2015HumanlevelCT} and robotics \citep{Levine2018LearningHC, Andrychowicz2020LearningDI}. Through interaction with an unknown environment, RL agents iteratively improve their policy by learning to maximize a reward signal, which has the potential to be used in lieu of hand-crafted control policies. However, the performance of policies learned by RL is found to be highly dependent on the careful specification of task-specific reward functions and, as a result, crafting a good reward function may require significant domain knowledge and technical expertise.

As an alternative to manual design of reward functions, \textit{inverse RL} (IRL) has emerged as a promising paradigm for policy learning. By framing the reward specification as a learning problem, operators can specify a reward function based on video examples. While \textit{imitation learning} typically requires demonstrations from a first-person perspective, IRL can in principle learn a reward function, \textit{i.e.}, a measure of task progression, from \textit{any} perspective, including third-person videos of humans performing a task. This has positive implications for data collection, since it is often far easier for humans to capture demonstrations in third-person.

Although IRL from third-person videos is appealing because of its perceived flexibility, learning a good reward function from raw video data comes with a variety of challenges. This is perhaps unsurprising, considering the visual and functional diversity that such data contains. For example, the task of pushing an object across a table may require different motions depending on the embodiment of the agent. A recent method for cross-embodiment IRL, dubbed XIRL \citep{zakka2022xirl}, learns to capture task progression from videos in a self-supervised manner by enforcing temporal cycle-consistency constraints. While XIRL can in principle consume any video demonstration, we observe that its ability to learn task progression degrades substantially when the visual appearance of the video demonstrations do not match that of the target environment for RL. Therefore, it is natural to ask the question: \textit{can we learn to imitate others from (a limited number of) diverse third-person videos?}

In this work, we demonstrate that it is indeed possible. Our key insight is that, while videos may be of great visual diversity, their underlying scene structure and agent-object interactions can be abstracted via a graph representation. Specifically, instead of directly using images, we extract object bounding boxes from each frame using an off-the-shelf detector, and construct a graph abstraction where each object is represented as a node in the graph. Often -- in robotics tasks -- the spatial location of an object by itself may not convey the full picture of the task at hand. For instance, to understand a task like \textit{Peg in Box} (shown in Figure \ref{fig:teaser}), we need to also take into account how the agent \emph{interacts} with the object. Therefore, we propose to employ \emph{Interaction Networks} \citep{battaglia2016interaction} on our graph representation to explicitly model interactions between entities. To train our model, we follow \citep{zakka2022xirl, dwibedi2019temporal} and apply a temporal cycle consistency loss, which (in our framework) yields task-specific yet embodiment- and domain-agnostic feature representations.

We validate our method empirically on a set of simulated cross-domain cross-embodiment tasks from X-MAGICAL \citep{zakka2022xirl}, as well as three vision-based robotic manipulation tasks. To do so, we collect a diverse set of demonstrations that vary in visual appearance, embodiment, object categories, and scene configuration; X-MAGICAL demonstrations are collected in simulation, whereas our manipulation demonstrations consist of real-world videos of humans performing tasks. We find our method to outperform a set of strong baselines when learning from visually diverse demonstrations, while simultaneously matching their performance in absence of diversity. Further, we demonstrate that vision-based policies trained with our learned reward perform tasks with greater precision than human-designed reward functions, and successfully transfer to a real robot setup with only approximate correspondence to the simulation environment. Thus, our proposed framework completes the cycle of learning rewards from real-world human demonstrations, learning a policy in simulation using learned rewards, and finally deployment of the learned policy on physical hardware.

\vspace{-0.025in}
\section{Related Work}
\label{sec:related_work}
\vspace{-0.1in}
\textbf{Learning from demonstration.} Conventional imitation learning methods require access to expert demonstrations comprised of observations and corresponding ground-truth actions for every time step \citep{pomerleau1988bc, Atkeson1997RobotLF, argall2009lfd, Ravichandar2020Recent}, for which kinesthetic teaching or teleoperation are the primary modes of data collection in robotics. To scale up learning, video demonstrations are recorded with human operating the same gripper that the robot used, which also allows direct behaviro cloning~\cite{song2020grasping,young2020visual}. More recently, researchers have developed methods that instead infer actions from data via a learned forward \citep{Pathak2018ZeroShotVI} or inverse \citep{Torabi2018BehavioralCF, Radosavovic2021StateOnlyIL} dynamics model. However, this approach still makes the implicit assumption that imitator and demonstrator share a common observation and action space, and are therefore not directly applicable to the cross-domain cross-embodiment problem setting that we consider. 


\textbf{Inverse RL.} To address the aforementioned limitations, inverse RL has been proposed~\citep{Ng2000,Abbeel2004,ho2016generative,Fu2017,Aytar2018,Torabi2018g} and it has recently emerged as a promising paradigm for cross-embodiment imitation in particular~\citep{schmeckpeper2020reinforcement, jin2020visual, xiong2021learning, lee2021generalizable, chen2021learning, fickinger2021cross, qin2021dexmv, zakka2022xirl, arunachalam2022dexterous}. For example, \citet{schmeckpeper2020reinforcement} proposes a method for integrating video demonstrations without corresponding actions into off-policy RL algorithms via a latent inverse dynamics model and heuristic reward assignment, and \citet{zakka2022xirl} (XIRL) learns a reward function from video demonstrations using temporal cycle-consistency and trains an RL agent to maximize the learned rewards. In practice, however, inverse RL methods such as XIRL are found to require limited visual diversity in demonstrations. Our work extends XIRL to the setting of diverse videos by introducing a graph abstraction that models agent-object and object-object interactions while still enforcing temporal cycle-consistency. 




\textbf{Object-centric representations.} have been proposed in many forms at the intersection of computer vision and robotics. For example, object-centric scene graphs can be constructed for integrated task and motion planning \citep{Fainekos2009TemporalLM, srivastava2014combined, zhu2021hierarchical}, navigation \citep{Gupta2019CognitiveMA, Yang2019VisualSN}, relational inference \citep{Xu2017SceneGG, Li2017SceneGG}, dynamics modeling \citep{battaglia2016interaction, watters2017visual, materzynska2020something, ye2019compositional, qi2021learning}, model predictive control~\citep{sanchez2018graph,li2019learning,ye2020object} or visual imitation learning \citep{sieb2020graph}. Similar to our work, \citet{sieb2020graph} propose to abstract video demonstrations as object-centric graphs for the problem of single-video cross-embodiment imitation, and act by minimizing the difference between the demonstration graph and a graph constructed from observations captured at each step. As such, their method is limited to same-domain visual trajectory following, whereas we learn a general alignment function for cross-domain cross-embodiment imitation and leverage \textit{Interaction Networks} \citep{battaglia2016interaction} for modeling graph-abstracted spatial interactions rather than relying on heuristics.

 \begin{figure}
    \centering
    \includegraphics[width=0.95\textwidth]{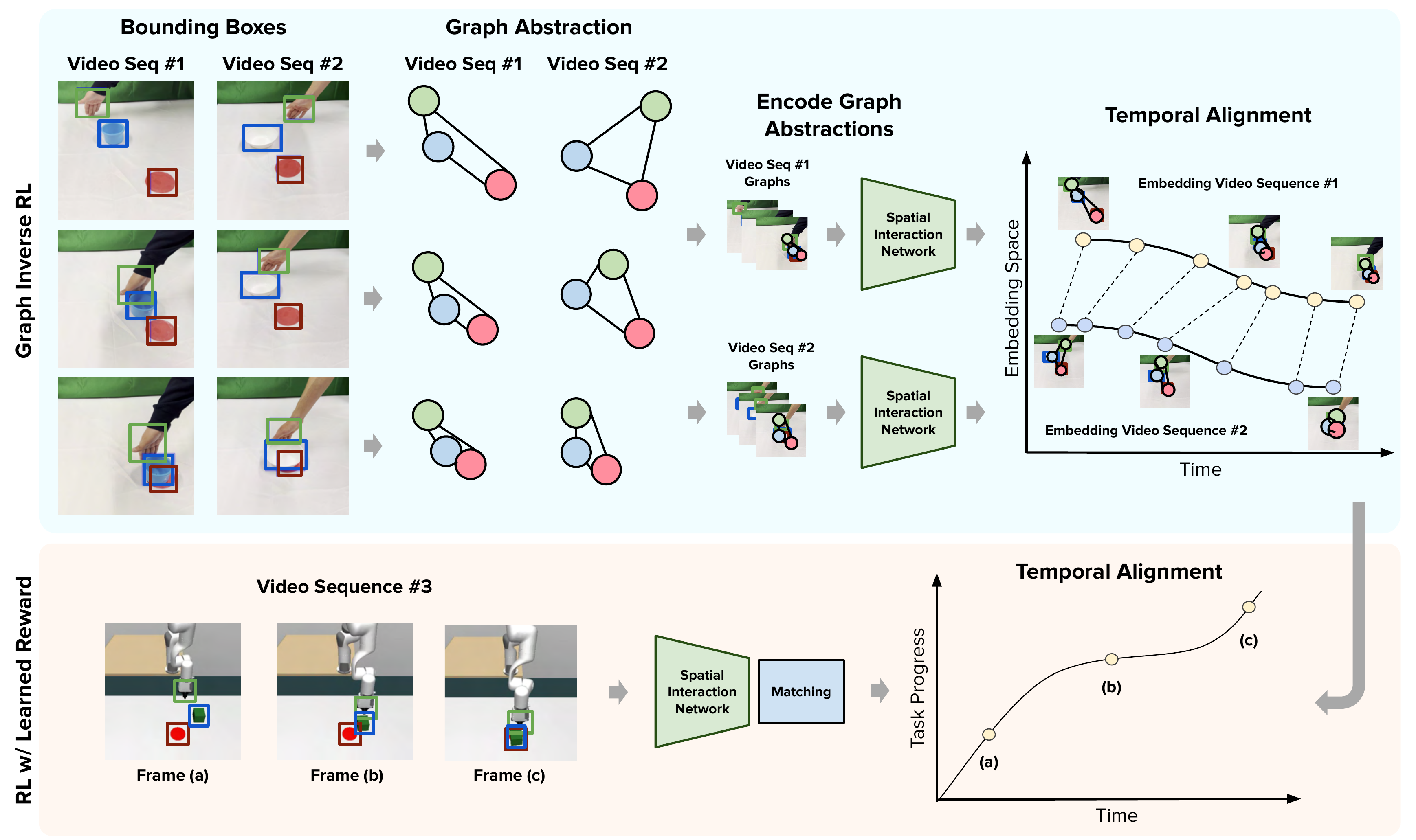}
    \vspace{-0.025in}
    \caption{\textbf{Overview.} We extract object bounding boxes from video sequences using an off-the-shelf detector, and construct a graph abstraction of the scene. We model graph-abstracted object interactions using \emph{Interaction Networks} \citep{battaglia2016interaction}, and learn a reward function by aligning video embeddings temporally. We then train image-based RL policies using our learned reward function, and deploy on a real robot.}
    \label{fig:approach}
    \vspace{-0.1in}
\end{figure}

\section{Our Approach}
\label{sec:approach}
\vspace{-0.075in}
In this section, we describe our main contribution, which is a self-supervised method for learning a reward function directly from a set of diverse third-person video demonstrations by applying temporal matching on graph abstractions. Our Graph Inverse Reinforcement Learning (GraphIRL) framework, shown in Figure \ref{fig:approach}, consists of building an object-centric graph abstraction of the video demonstrations and then learn an embedding space that captures task progression by exploiting the temporal cue in the videos. This embedding space is then used to construct a \emph{domain invariant} and \emph{embodiment invariant} reward function which can be used to train any standard reinforcement learning algorithm.

 \noindent {\textbf{Problem Formulation.}} Given a task $T$, our approach takes a dataset of video demonstrations $D = \{V_1, V_2, \dots, V_n\}$. Each video consists of image frames $\{I_1^i, I_2^i, \dots, I_k^i\}$  where $i$ denotes the video frame index and $k$ denotes the total number of frames in $V_i$. Given $D$, our goal is to learn a reward function that can be used to solve the task $T$ for any robotic environment. Notably, we do \emph{not} assume access to any action information of the expert demonstrations, and our approach does \emph{not} require objects or embodiments in the target environment to share appearance with demonstrations.

\subsection{Representation Learning}
\label{sec:rep_learning}
\vspace{-0.06in}
To learn task-specific representations in a self-supervised manner, we take inspiration from \citet{dwibedi2019temporal} and employ a temporal cycle consistency loss. However, instead of directly using images, we propose a novel object-centric graph representation, which allows us to learn an embedding space that not only captures task-specific features, but depends \emph{solely} on the spatial configuration of objects and their interactions. We here detail each component of our approach to representation learning.

\noindent {\textbf{Object-Centric Representation.}} Given video frames $\{I_1^i, I_2^i, \dots, I_k^i\}$, we first extract object bounding boxes from each frame using an off-the-shelf detector. Given $N$ bounding boxes for an image, we represent each bounding box as a $4 + m$ dimensional vector $o_j = \{x_1, y_1, x_2, y_2, d_1, d_2, \dots, d_m\}$, where the first $4$ dimensions represent the leftmost and rightmost corners of the bounding box, and the remaining $m$ dimensions encode distances between the centroids of the objects.
For each frame $I^i_j$ we extract an object-centric representation $I'^{i}_j = \{ o_1, o_2, \dots, o_m\}$ such that we can represent our dataset of demonstrations as $D' = \{V'_1, V'_2, \dots, V'_n\}$ where $V'_i$ is the sequence of bounding boxes corresponding to video $V_i$. Subsequent sections describe how we learn representations given $D'$.

\noindent {\textbf{Spatial Interaction Encoder.}} Taking inspiration from recent approaches on modeling physical object dynamics \citep{battaglia2016interaction, watters2017visual}, we propose a \emph{Spatial Interaction Encoder Network} to explicitly model object-object interactions. Specifically, given a sequence $V'_i$ from $D'$, we model each element $I'$ as a graph, $G = (O, R)$, where $O$ is the set of objects $\{o_1, o_2, \dots, o_m\}$, $m$ is the total number of objects in $I'$, and $R$ denotes the relationship between objects (\emph{i.e.}, whether two objects interact with each other). For simplicity, all objects are connected with all other objects in the graph such that $R= \{(i,j)  \mid i \ne j  \land i \leq m \land j \leq m \}$. We compose an object embedding for each of $o_i \in O$ by combining \emph{self} and \emph{interactional} representations as follows:
\begin{align}
\label{eq:sie_self}
    f_{o}(o_i) = \phi_{\text{agg}} (f_{\text{s}} + f_{\text{in}}) \quad\text{with}\quad f_s(o_i) = \phi_s(o)\,, \quad f_{\text{in}}(o_i) = \sum_{j=1}^{m} \phi_{\text{in}}((o_i, o_j)) \mid (i, j) \in \mathbb{R}\,,
\end{align}
where $f_s(o_i)$ represents the \emph{self} or independent representation of an object, $f_{\text{in}}$ represents the \emph{interactional} representation, \emph{i.e.}, how it interacts with other objects in the scene, $f_{o}$ is the final object embedding, and $(,)$ represents concatenation. Here, the encoders $\phi_\text{s}$, $\phi_{\text{in}}$ and $\phi_{\text{agg}}$ denote Multi layer Perceptron (MLP) networks respectively. We emphasize that the expression for $f_\text{in}(\cdot)$ implies that the object embedding $f_o(.)$ depends on \emph{all} other objects in the scene; this term allows us to model relationships of an object with the others. The final output from the spatial interaction encoder $\psi(\cdot)$ for object representation $I'$ is the mean of all object encodings:
\begin{align}
\label{eq:sie_final_eq}
    \psi(I') = \frac{1}{m} \sum_i^{m} f(o_i)\,.
\end{align}
\label{approach:sie}
The spatial interaction encoder is then optimized using the temporal alignment loss introduced next.

\noindent {\textbf{Temporal Alignment Loss}}. Taking inspiration from prior works on video representation learning \citep{dwibedi2019temporal, haresh2021learning, liu2021learning, wang2020dynamic, hadji2021representation}, we employ the task of temporal alignment for learning task-specific representations. Given a pair of videos, the task of self-supervised alignment implicitly assumes that there exists true semantic correspondence between the two sequences, \textit{i.e.}, both videos share a common semantic space. These works have shown that optimizing for alignment leads to representations that could be used for tasks that require understanding task progression such as action-classification. This is because in order to solve for alignment, a learning model has to learn features that are (1) common across most videos and (2) exhibit temporal ordering.  For a sufficiently large dataset with single task, the most common visual features would be distinct phases of a task that appear in all videos and if the task has small permutations, these distinct features would also exhibit temporal order. In such scenarios, the representations learned by optimizing for alignment are \emph{task-specific} and invariant to changes in viewpoints, appearances and actor embodiments.   

In this work, we employ Temporal Cycle Consistency (TCC) \citep{dwibedi2019temporal} loss to learn temporal alignment. TCC optimizes for alignment by learning an embedding space that maximizes one-to-one nearest neighbour mappings between sequences. This is achieved through a loss that maximizes for cycle-consistent nearest neighbours given a pair of video sequences. In our case, the cycle consistency is applied on the \emph{graph abstraction} instead of image features as done in the aforementioned video alignment methods. Specifically, given $D'$, we sample a pair of bounding box sequences $V'_i = \{I'^{i}_1, \dots, I'^{i}_{m_i} \}$ and $V'_j = \{I'{j}_1, \dots, I'^{j}_{m_j} \}$ and extract embeddings by applying  the spatial interaction encoder defined in Equation \ref{eq:sie_final_eq}. Thus, we obtain the encoded features $S_i = \{\psi( I'^{i}_1), \dots, \psi (I'^{i}_{m_i})\}$ and $S_j  = \{\psi( I'^{j}_1), \dots, \psi (I'^{j}_{m_j})\}$. For the $n$th element in $S_i$, we first compute its nearest neighbour, $\upsilon^n_{ij}$, in $S_j$ and then compute the probability that it cycles-back to the $k$th frame in $S_i$ as:
\begin{align}
\label{eq:tcc_definition}
     \beta^k_{ijn} = \frac{e^{-||\upsilon ^n _{ij} - S_i^k ||^2}}{{\sum_k^{m_j} e^ {-|| \upsilon^n_{ij} - S^k_i || ^2}}}\,,
        \upsilon^n_{ij} = \sum_k^{m_j} \alpha_k s^k_j\,,
        \alpha_k = \frac{e^{-|| S^n_i - S^k_j  ||^2}}{{\sum_k^{m_j} e^ {-|| S^n_i - S^k_j || ^2}}}\,. 
\end{align}
The cycle consistency loss for $n$th element can be computed as  $L_n^{ij} = (\mu^n_{ij} - n)^2$, where  $\mu^n_{ij} = \sum^{mi}_k \beta^k_{ijn} k $ is the expected value of frame index $n$ as we cycle back. The overall TCC loss is then defined by summing over all pairs of sequence embeddings $(S_i, S_j)$ in the data, \textit{i.e.}, $L^n_{ij} = \sum_{ijn} L^n_{ij}$.



\subsection{Reinforcement Learning}
\vspace{-0.05in}
We learn a task-specific embedding space by optimizing for temporal alignment. In this section, we define how to go from this embedding space to a reward function that measures task progression. For constructing the reward function, we leverage the insight from \citet{zakka2022xirl} that in a task-specific embedding space, we can use euclidean distance as a notion of task progression, \textit{i.e.}, frames far apart in the embedding space will be far apart in terms of task progression and vice versa. We therefore choose to define our reward function as
\begin{align}
\label{eq:reward_function}
     r(o) = -\frac{1}{c} ||\psi(o) - g||^2\,, \quad\text{with}\quad g = \sum_{i=1}^n  \psi (I'^{i}_{m_{i}})\,,
\end{align}
where $o$ is the current observation, $\psi$ is the Spatial Interaction Encoder Network from Section \ref{sec:approach}, $g$ is the representative goal frame, $m_i$ is the length of sequence $V'^i$ and $c$ is a scaling factor. The scaling factor \emph{c} is computed as the average distance between the first and final observation of all the training videos in the learned embedding space. Note, that the range of the learned reward is $(-\infty, 0]$. Defining the reward function in this way gives us a dense reward because as the observed state gets closer and closer to the goal, the reward starts going down and approaches zero when the goal and current observation are close in embedding space. After constructing the learned reward, we can use it to train  any standard RL algorithm. We note that, unlike previous approaches \citep{schmeckpeper2020reinforcement, zakka2022xirl}, our method does not use \emph{any} environment reward to improve performance, and instead relies \emph{solely} on the learned reward, which our experiments demonstrate is sufficient for solving diverse robotic manipulation tasks.

\section{Experiments}
\label{sec:experiments}

In this section, we  demonstrate how our approach uses diverse video demonstrations to learn a reward function that generalizes to unseen domains. In particular,  we are interested in answering the questions: (1) How do vision-based methods for IRL perform when learning from demonstrations that exhibit \emph{domain shift}? and (2) is our approach capable of learning a stronger reward signal under this challenging setting? To that end, we first conduct experiments X-MAGICAL benchmark \citep{zakka2022xirl}. We then evaluate our approach on multiple robot manipulation tasks using a diverse set of demonstrations. 

\noindent{\textbf{Implementation Details.}} All MLPs defined in Equation \ref{eq:sie_final_eq} have 2 layers followed by a ReLU activation, and the embedding layer outputs features of size $128$ in all experiments. For training, we use ADAM~\citep{kingma2014adam} optimizer with a learning rate of $10^{-5}$. We use Soft Actor-Critic (SAC) \citep{haarnoja2018soft} as backbone RL algorithm for all methods. For experiments on X-MAGICAL, we follow \citet{zakka2022xirl} and learn a state-based policy; RL training is performed for $500$k steps for all embodiments. For robotic manipulation experiments, we learn a multi-view image-based SAC policy \citep{jangir2022look}. We train RL agent for $300$k, $800$k and $700$k steps for \emph{Reach}, \emph{Push} and \emph{Peg in Box} respectively. For fair comparison, we only change the learned reward function across methods and keep the RL setup identical. The success rates presented for all our experiments are averaged over $50$ episodes. Refer to Appendix \ref{sec:implementation_details} for further implementation details.

\textbf{\noindent{\textbf{Baselines.}}} We compare against multiple vision-based approaches that learn rewards in a self-supervised manner: \textbf{(1) {XIRL}}~\citep{zakka2022xirl} that learns a reward function by applying the TCC~\citep{dwibedi2019temporal} loss on demonstration video sequences, \textbf{(2) {TCN}}~\citep{sermanet2018time} which is a self-supervised contrastive method for video representation learning that optimizes for temporally disentangled representations, and \textbf{(3) LIFS}~\citep{gupta2017learning} that learns an invariant feature space using a dynamic time warping-based contrastive loss. Lastly, we also compare against the manually designed \textbf{(4) Environment Rewards} from \citet{jangir2022look}. For vision-based baselines, we use a ResNet-18 encoder pretrained on ImageNet \citep{ILSVRC15} classification.  We use the hyperparameters, data augmentation schemes and network architectures provided in \citet{zakka2022xirl} for all vision-based baselines. Please refer to Appendix \ref{sec:env_reward_desc}
for description of environment rewards and \citet{zakka2022xirl} for details on the vision-based baselines.

\input{tables/xmagical/overview-long}

\begin{figure}[t]
    \centering
    \includegraphics[width=1\textwidth]{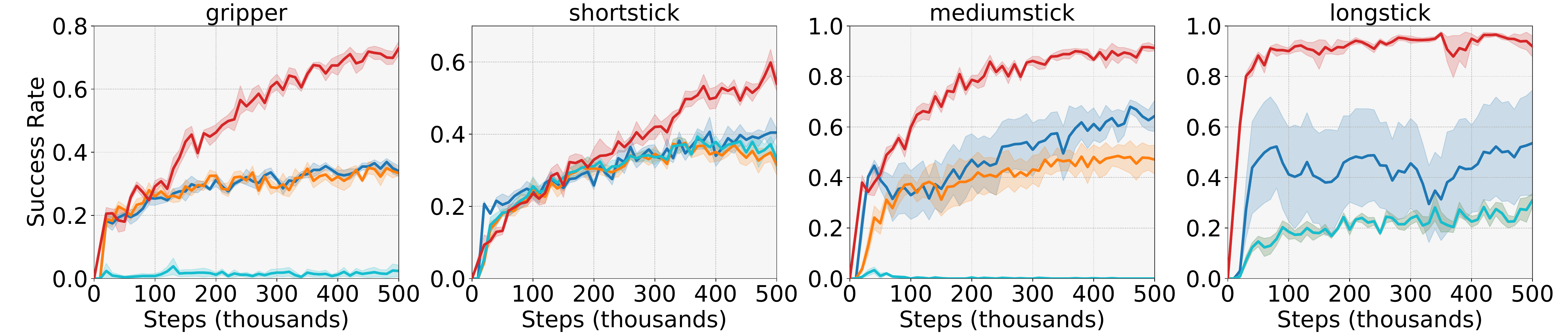}\vspace{0.025in}\\
    \includegraphics[width=1\textwidth]{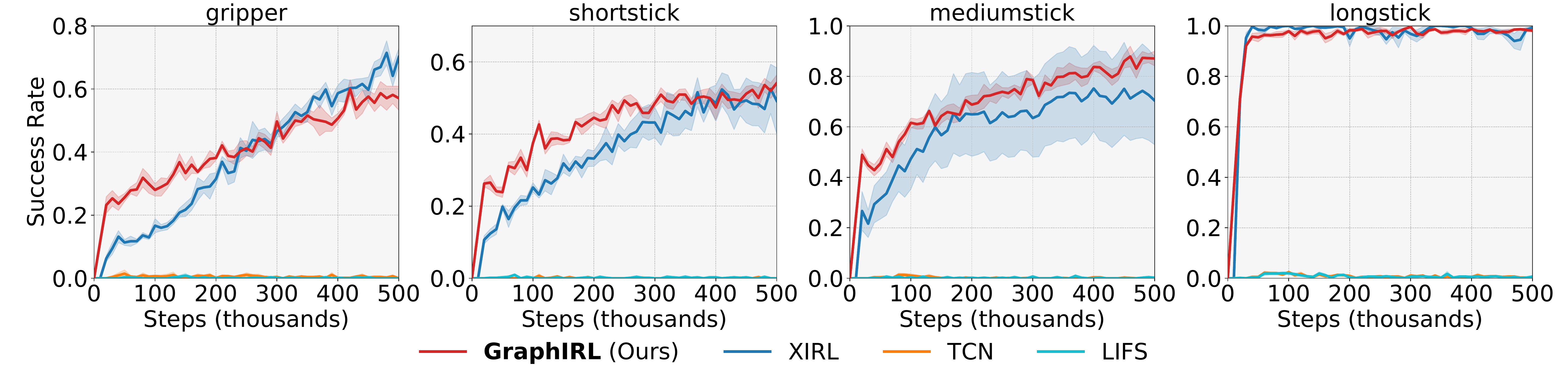}
    \vspace{-0.2in}
    \caption{\textbf{Cross-Embodiment Cross-Environment.} Success rates of our method \emph{GraphIRL} and baselines on \textit{(top)} Standard Environment Pretraining $\rightarrow$ Diverse Environment RL and \textit{(bottom)} Diverse Environment Pretraining $\rightarrow$ Standard Environment RL. All reported numbers are averaged over 5 seeds. Our approach performs favorably when compared to other baselines on both settings.}  
    \label{fig:cross-env-cross-emb-standard2diverse}
     \vspace{-0.15in}
\end{figure}

\subsection{Experimental Setup}
\label{sec:setup}
We conduct experiments under two settings: the \emph{Sweep-to-Goal} task from X-MAGICAL \citep{zakka2022xirl}, and robotic manipulation tasks with an xArm robot both in simulation and on a real robot setup. We describe our experimental setup under these two settings in the following.

\noindent \textbf{X-MAGICAL.} We choose to extend X-MAGICAL \citep{zakka2022xirl}, a 2D simulation environment for cross-embodiment imitation learning. On this benchmark, we consider a multi-object sweeping task, where the agent must push three objects towards a static goal region. We utilize two variants of the X-MAGICAL benchmark, which we denote as \textit{\textbf{Standard}} (original) and \textit{\textbf{Diverse}} (ours) environments, shown in Figure \ref{fig:xmagical-variants}. \textit{\textbf{Standard}} only randomizes the position of objects, whereas \textit{\textbf{Diverse}} also randomizes visual appearance. We consider a set of four unique embodiments $\{$\textit{gripper, short-stick, medium-stick, long-stick}$\}$. In particular, we conduct experiments in the \emph{cross-environment} and \emph{cross-embodiment setting} where we learn a reward function in the \textbf{\textit{Standard}} environment on 3 held-out embodiments and do RL in the \textit{\textbf{Diverse}} environment on 1 target embodiment, or vice-versa. This provides an additional layer of difficulty for the RL agent as visual randomizations show the brittleness of vision-based IRL methods. Refer to Appendix \ref{sec:X_Magical_details} for more details on performed randomizations.


\noindent \textbf{Robotic Manipulation.} Figure \ref{fig:teaser} shows initial and success configurations for each of the three task that we consider: \textbf{(1) Reach} in which the agent needs to reach a goal (red disc) with its end-effector, \textbf{(2) Push} in which the goal is to push a cube to a goal position, and \textbf{(3) Peg in Box} where the goal is to put a peg tied to the robot's end-effector inside a box. The last task is particularly difficult because it requires geometric 3D understanding of the objects. Further, a very specific trajectory is required to avoid collision with the box and complete the task. We collect a total of $256$ and $162$ video demonstrations for \emph{Reach} and \emph{Peg in Box}, respectively, and use $198$ videos provided from \citet{schmeckpeper2020reinforcement} for \emph{Push}. The videos consist of human actors performing the same tasks but with a number of diverse objects and goal markers, as well as varied positions of objects. Unlike the data collected by \citet{schmeckpeper2020reinforcement}, we do not fix the goal position in our demonstrations. In order to detect objects in our training demonstrations, we use a trained model from \citet{shan2020understanding}. The model is trained on a large-scale dataset collected from YouTube and can detect hands and objects in an image.; refer to Appendix \ref{sec:xArm_data_details} for more details on data collection. Additionally, we do not require the demonstrations to resemble the robotic environment in terms of appearance or distribution of goal location. We use an xArm robot as our robot platform and capture image observations using a static third-person RGB camera in our real setup; details in Appendix \ref{sec:robot_setup}. 

\vspace{-0.05in}
\subsection{Results}
\label{sec:results}
\textbf{X-MAGICAL.} Results for the \emph{cross-embodiment and cross-environment} setting are shown in Figure \ref{fig:cross-env-cross-emb-standard2diverse}. When trained on \textit{Standard}, our method performs significantly better than vision-based baselines (\textit{e.g.}, $0.58$ GraphIRL for gripper vs $0.35$ for XIRL and $0.99$ GraphIRL for longstick vs $0.56$ XIRL). We conjecture that vision-based baselines struggle with visual variations in the environment, which our method is unaffected by due to its graph abstraction. Additionally, when trained on \emph{diverse} environment, GraphIRL outperforms $3$ out of $4$ embodiments.
\begin{figure}[t]
    \centering
    \includegraphics[width=0.8\textwidth]{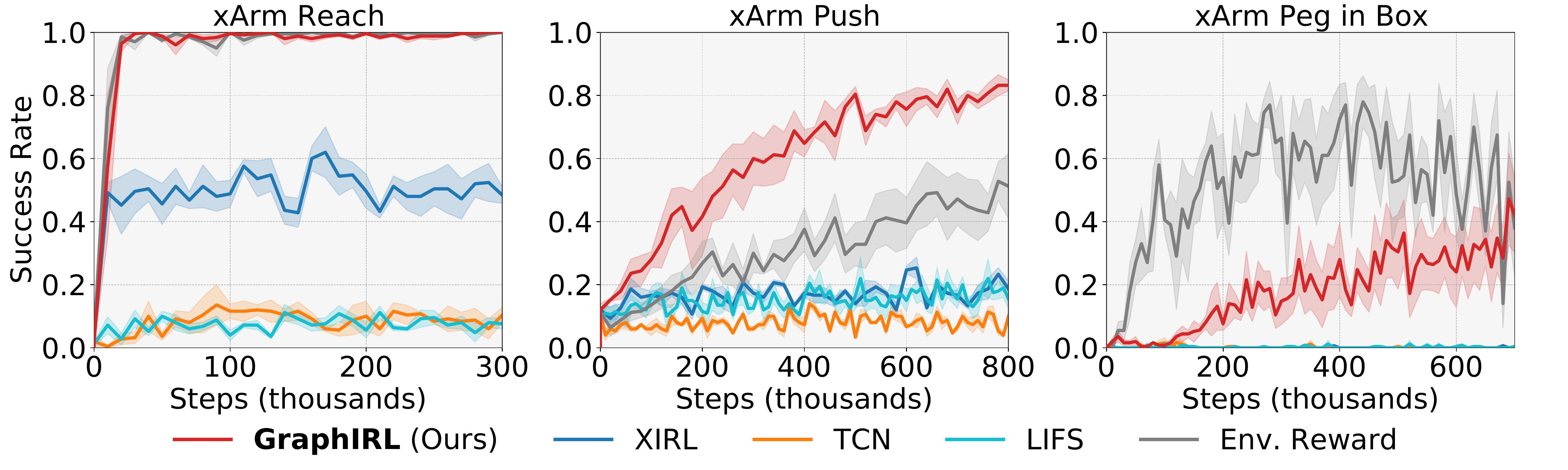}
    \caption{\textbf{Robotic Manipulation.} Success rates of our method \emph{GraphIRL} and baselines on the tasks of \emph{Reach}, \emph{Push} and \emph{Peg in Box}. All results are averaged over 5 seeds. We observe significant gains in performance specially over vision-based baselines due to large \emph{domain-gap}}
    \label{fig:xarm-tasks}
\end{figure}

\input{tables/robot-real}

\textbf{Robotic manipulation in simulation.} In this section, we answer the core question of our work: \emph{can we learn to imitate others from diverse third-person videos?} In particular, we collect human demonstrations for manipulation tasks as explained in Section \ref{sec:setup} and learn a reward function as explained in Section \ref{sec:approach}. This is a challenging setting because as shown in Figure \ref{fig:teaser}, the collected data and robotic environments belong to different domains and do not share any appearance characteristics. Further, unlike previous works \citep{schmeckpeper2020reinforcement, zakka2022xirl}, we do not use any environment reward as an additional supervision to the reinforcement learning agent. Figure \ref{fig:xarm-tasks} presents our results. For the \textbf{Reach} task, GraphIRL and environment reward are able to achieve a success rate of $1.0$, while other baseline methods are substantially behind GraphIRL (e.g. $0.477$ XIRL and $0.155$ TCN). The poor performance of vision-based baselines could be attributed to substantial visual domain shift. Due to domain shift, the learned rewards for these baselines produce low rewards for successful episodes, please refer to Appendix \ref{sec:qualitative_analysis} for a more detailed qualitative analysis.  In the \textbf{Push} setting, we find that vision-based baseline methods still perform poorly. Similar to \textbf{Reach}, XIRL performs the best out of the vision-based baselines with a success rate of $0.187$, and GraphIRL performs better than environment reward (e.g. $0.832$ GraphIRL vs $0.512$ Environment Reward). This result shows clear advantage of our method as we are able to outperform a hand-designed reward function without using any task specific information. The \textbf{Peg in Box} task is rigorous to solve since it requires 3-d reasoning and a precise reward function. Here, while all vision-based methods fail, our GraphIRL method is able to solve the task with a success rate comparable to that achieved with the hand-designed environment reward. Overall, our GraphIRL method is able to solve 2D and 3D reasoning tasks with a real-robot without a hand-designed reward function or access to 3D scene information.

\textbf{Real robot experiments.} Finally, we deploy the learned policies on a real robot. For each experiment, we conduct $15$ trials per method and report the average success rate. Results are shown in Table \ref{tab:robot_real}. Interestingly, we find that GraphIRL outperforms XIRL in all three tasks on the real robot setup (e.g. $0.26$ XIRL vs $0.86$ GraphIRL on \emph{Reach} and  $0.27$ XIRL vs $0.60$ GraphIRL on \emph{Push}), and on \emph{Push}, GraphIRL performs better than the environment reward specifically designed for the task (e.g. $0.47$ Environment Reward vs $0.6$ GraphIRL) which is in line with our findings in simulation.

\input{tables/ablation}

\subsection{Ablations}

\label{sec:ablations}
In this section, we perform ablation study using the \emph{Push} task to validate our design choices in Section \ref{sec:approach}. In the experiments below, we perform RL training for $500$k steps and report the final success rate.

\textbf{Impact of Modelling Spatial Interactions.} We study the impact of modeling object-object spatial interactions using Spatial Interaction Encoder Network described (IN) in Section \ref{sec:rep_learning}. Specifically, we replace our proposed encoder component with an Multi-Layer Perceptron (MLP) by concatenating representations of all objects into a single vector and then feeding it to a 3-layer MLP network. As shown in Table \ref{tab:ablation_interactions}, IN leads to a $20\%$ improvement in the reinforcement learning success rate. 

\textbf{Impact of Decreasing Number of Demonstration Videos.} 
As shown in Table \ref{tab:ablation_data}, the performance of our approach gradually decreases as we decrease demonstration data. However, we note that GraphIRL achieves $67\%$ success rate with $25$\% of total training videos (49 videos). 
This demonstrates that our approach is capable of learning meaningful rewards even with a small number of videos.

\section{Conclusions and Limitations}
\label{sec:conclusion}
We demonstrate the effectiveness of our proposed method, \textit{GraphIRL}, in a number of IRL settings with diverse third-person demonstrations. In particular, we show that our method successfully learns reward functions from human demonstrations with diverse objects and scene configurations, that we are able to train image-based policies in simulation using our learned rewards, and that policies trained with our learned rewards are more successful than both prior work and manually designed reward functions on a real robot.  With respect to limitations, while our method relaxes the requirements for human demonstrations, collecting the demonstrations still requires human labor; and although our results indicate that we can learn from relatively few videos, eliminating human labor entirely remains an open problem.

\clearpage


\normalsize{{\bibliography{main.bib}}}

\setcounter{section}{0}

\renewcommand\thesection{\Alph{section}}
\begin{appendices}

\section{Qualitative Analysis of Learned Reward}

\label{sec:qualitative_analysis}
In this section, we present qualitative analysis of the reward learned using \emph{GraphIRL}. We plot the reward as defined in Equation \ref{eq:reward_function} for \textit{GraphIRL} and two baseline IRL methods for three test examples across three tasks. The tasks we evaluate with are \textit{Peg in Box}, \textit{Push}, and \textit{Reach}. For each task, we use show two successful episodes and one unsuccessful episode. The length of each episode is 50, and for each figure we have included, we provide images that align with critical points in the completion of the task. 


\input{figures/appendix-A/pegbox/2/pegbox_success}
\input{figures/appendix-A/pegbox/0/pegbox_failure}
\newpage
\input{figures/appendix-A/push/2/push_success}
\input{figures/appendix-A/push/0/push_failure}
\newpage
\input{figures/appendix-A/reach/2/reach_success}
\input{figures/appendix-A/reach/0/reach_failure}

\normalsize{We find that our method provides a superior and accurate reward signal to the agent compared to the baseline visual IRL methods. We observe that if a task is being completed successfully or unsuccessfully in a video, our method can obtain a reward that accurately reflects how close the agent is to completing the task. Additionally, both \emph{XIRL} and \emph{TCN} yield low reward even for successful episodes due to large distance between the current observation and the representative goal observation in the embedding space which could be attributed to visual domain shift.}


\section{Additional Implementation Details}
\label{sec:implementation_details}
\input{tables/hyperparameters}
\normalsize{\noindent{\textbf{Representation Learning.}}} Each MLP in the Spatial Interaction Encoder Network defined in Equation \ref{approach:sie} is implemented as a 2-layer network with a ReLU activation. The size of the final embedding $\psi(\cdot)$ is $128$ in our experiments. Please see Table \ref{tab:hyperparameters} for a detailed list of hyperparameters for representation learning. All the hyperparameters in Table \ref{tab:hyperparameters} are kept fixed for all tasks considered in this work.

\noindent{\textbf{Reinforcement Learning.}} For X-MAGICAL, we follow \citet{zakka2022xirl} and learn a state based policy. The state vector has dimensions of 16 and 17 for the \emph{Standard} and \emph{Diverse} environments respectively. The \emph{Diverse} environment state has an additional dimension to represent the size of blocks. For xArm, we learn an image based policy. Specifically, we use first-person and third-person cameras to learn a policy from multi-view image data. We extract $84 \times 84$ image from both cameras and concatenate them channel-wise. We use the network architecture and attention mechanism proposed in \citet{jangir2022look}. Additionally, we apply data augmentation techniques: random $\pm4$ pixel shift \citep{yarats2020image} and color jitter \citep{hansen2021stabilizing}.

\noindent{\textbf{Extracting Reward.}} In order to compute the reward during Reinforcement Learning (RL) training, we use the locations of objects available in simulation to extract the bounding boxes corresponding to the current observation. The bounding boxes are used to construct the object representation which is then passed to the trained Spatial Interaction Encoder Network to get the reward.


\noindent{{\textbf{Criterion for Success.}}} We use distance threshold to determine the success of an episode. The thresholds are $5$cms, $10$cms and $8$cms for \emph{Reach}, \emph{Push} and \emph{Peg in Box} respectively. The distance refers to distance between goal position and end-effector for \emph{Reach}, and goal position and object position for \emph{Push} and \emph{Peg in Box}.


\noindent{\textbf{Baseline Implementation Details.}} For all the vision-based baselines, we use the hyperparameters, data augmentation schemes and network architectures provided in \citet{zakka2022xirl}. Readers are encouraged to read \citet{zakka2022xirl} for more details on the vision-based baselines.

\section{X-MAGICAL Experiment Details}
\label{sec:X_Magical_details}
\subsection{Demonstration Data}

For collecting demonstration data in the X-MAGICAL \textit{Diverse} environment, we trained 5 uniquely-seeded Soft Actor-Critic (SAC) RL policies for $2$ million steps for each embodiment using the environment reward. We collect $1000$ successful episode rollouts for each embodiment using the 5 trained policies. In particular, each policy is used to produce $200$ episode rollouts for a given embodiment.

\subsection{Diverse Environment}

Below, we explain the randomization performed on the blocks in the diverse environment that we use in our experiments:

\begin{itemize}
    \item \textbf{Color:} We randomly assign 1 out of 4 colors to each block.
    
    \item \textbf{Shape:} Each block is randomly assigned 1 out of 6 shapes.
    
    \item \textbf{Size:} The block sizes are also varied. In particular, we generate a number between $0.75$ and $1.25$ and multiply the default block size by that factor.
    
    \item \textbf{Initial Orientation:} The initial orientation of the blocks is also randomized. We randomly pick a value between $0$ to $360$ degrees.
    
    \item \textbf{Initial Location:} The initial location of the boxes is randomized by first randomly picking a position for the y-coordinate for all blocks and then randomly selecting x-coordinate separately for each block. This randomization is also performed in the standard environment.
\end{itemize}

\section{Additional Results on X-MAGICAL Benchmark}
\begin{figure}[ht!]
    \centering
    \includegraphics[width=1\textwidth]{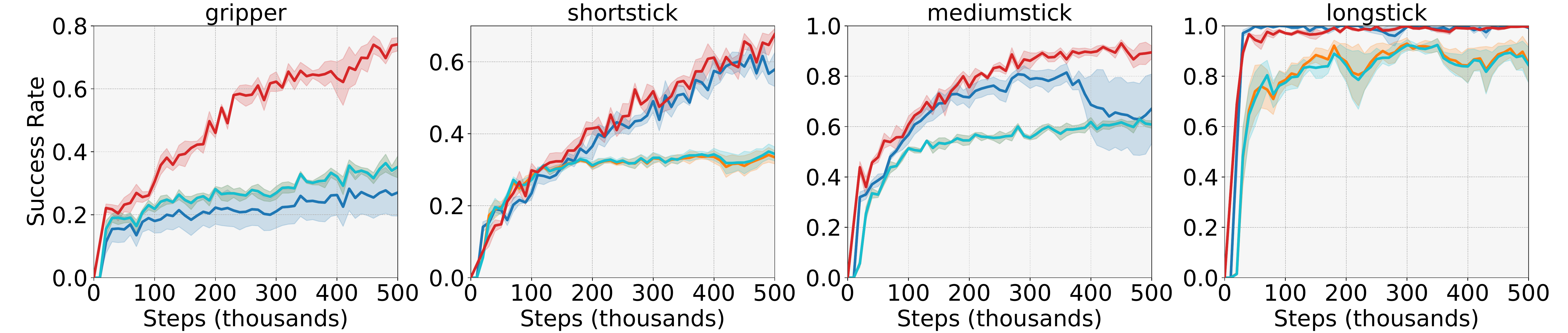}
    \includegraphics[width=1\textwidth]{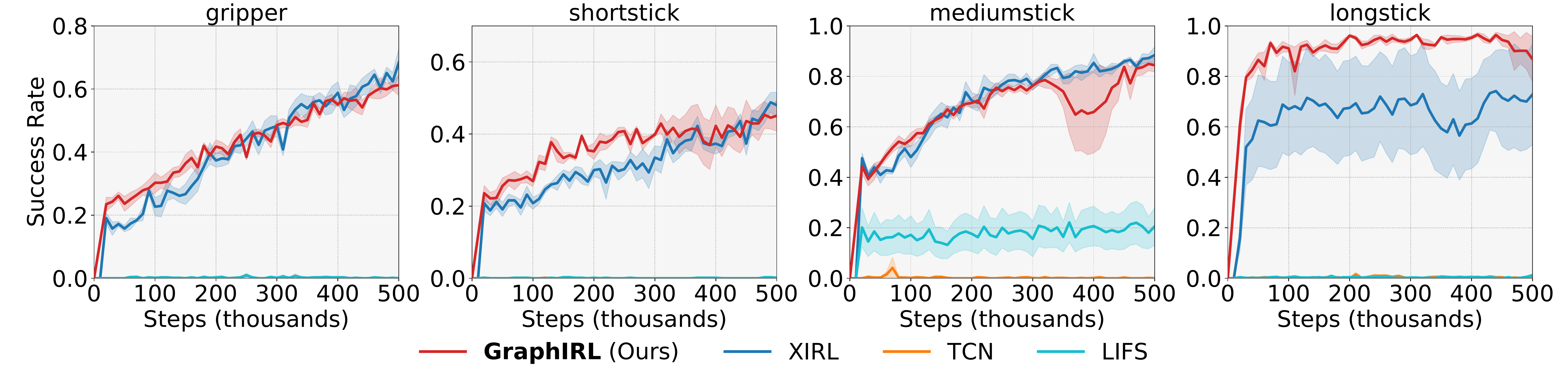}
    \caption{\textit{\textbf{Cross-Embodiment Same-Environment:}} We further evaluate \textit{GraphIRL} in the cross-embodiment same-environment setting \textit{(top)} Standard Environment \textit{(bottom)} Diverse Environment, and it continues to provide competitive success rates akin to those achieved by XIRL. These results confirm that \textit{GraphIRL} is a consistent and reliable method for learning from video demonstrations in visually similar environments.}
    \label{fig:same-env-cross-emb}
\end{figure}
\normalsize{To complement our cross-embodiment cross-environment results from the main paper, we also report results for \textit{X-MAGICAL} in the \emph{cross-embodiment} \emph{same-environment} setting. As shown in Figure \ref{fig:same-env-cross-emb}, we outperform \emph{TCN} and \emph{LIFS} by significant margins and achieve comparable results to \emph{XIRL}. These results reflect the effectiveness of \textit{GraphIRL} when learning in a visually similar environment with visually different agents.}

\section{Appendix E: xArm Experiment Details}
\label{sec:xArm_details}
\subsection{Description of Environment Rewards}

\label{sec:env_reward_desc}
In this section, we define the environment rewards for xArm environments that were compared against GraphIRL in robot manipulation experiments under Section \ref{sec:experiments}. We define $p_{g}$, $p_{o}$, and $p_{e}$ as the positions of the goal, object and robot end-effector respectively. The reward for \emph{Push} is defined as $||p_o - p_g ||^2$, for reach it becomes $||p_e - p_g||^2$ and finally for \emph{Peg in Box}, the reward is $|| p_o - p_g||^2$. Note that the distances are computed using 2-d positions in the case of \emph{Reach} and \emph{Push} and 3-d positions in the case of \emph{Peg in Box}. 

\subsection{Demonstration Data}
\label{sec:xArm_data_details}
We use data from \citep{schmeckpeper2020reinforcement} for \emph{Push}. We collect $256$ and $162$ demonstrations respectively for \emph{Reach} and \emph{Peg in Box}. For \emph{Reach}, we use 18 visually distinct goal position markers \textit{i.e.} $3$ different shapes and each shape with $6$ different colors in order to ensure visual diversity. \emph{Reach} demonstrations have minimum, average and maximum demonstration lengths of $1.76$ seconds, $4.51$ seconds and $9.23$ seconds respectively. For \emph{Peg in Box}, we use 4 visually distinct objects. In this case, the minimum, average and maximum demonstration lengths are $1.73$ seconds, $4.74$ seconds and $11.7$ seconds respectively. For both \emph{Reach} and \emph{Peg in Box}, the goal and object positions are also varied across demonstrations to diversify trajectories. Please see \url{https://sateeshkumar21.github.io/GraphIRL/} for examples of collected demonstrations.


\section{Additional Results on Robot Manipulation in Simulation}
\begin{figure}[ht!]
    \centering
    \includegraphics[width=0.7\textwidth]{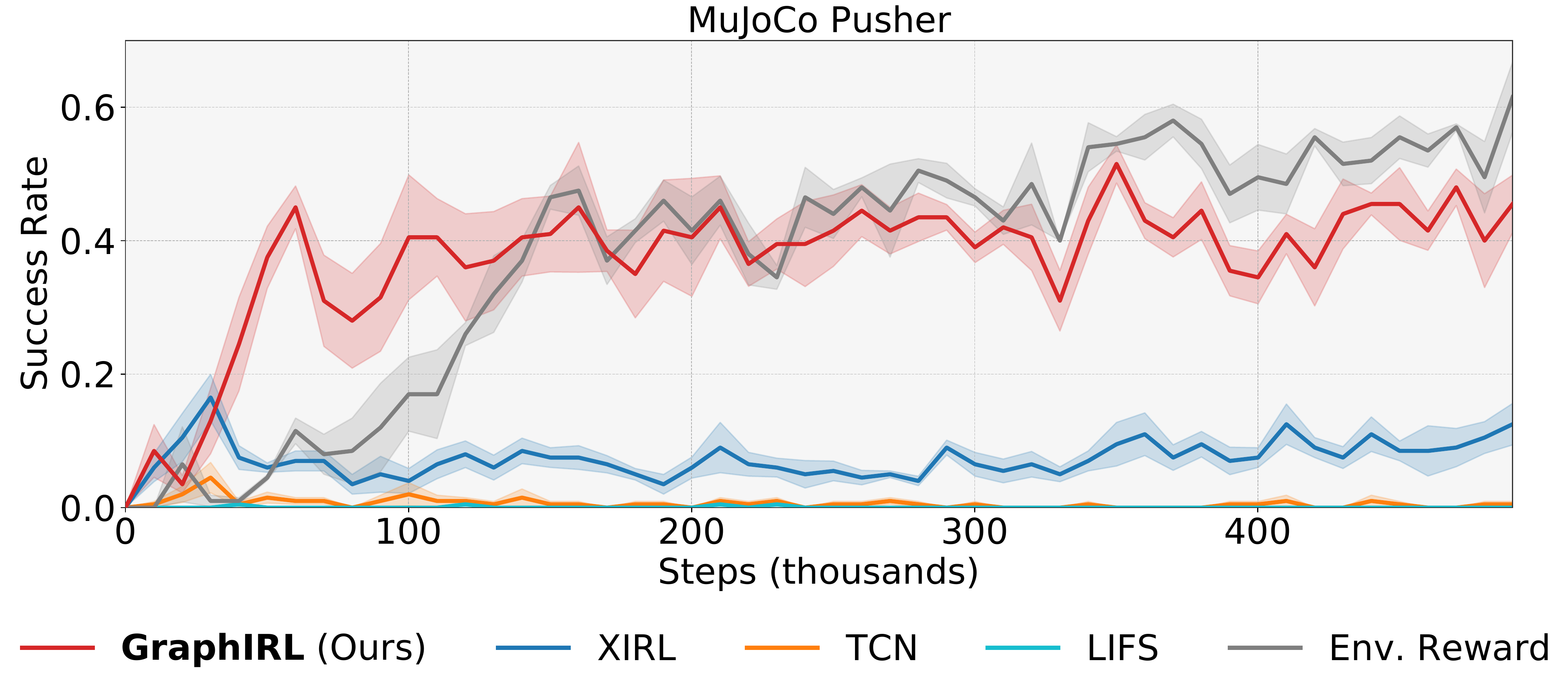}
    \caption{\textbf{MuJoCo State Pusher Task Progress: Success} \textit{GraphIRL} provides a reward signal that is both better than all other vision-based baselines and nearly as good as the task-specific environment reward. This indicates that the reward learned from \textit{GraphIRL} could be used across multiple environments of the same task, showing strong generalization capabilities.}
    \label{fig:mujoco_pusher}
\end{figure}

\normalsize{We also experiment with the \emph{MuJoCo State Pusher} environment used by \citet{schmeckpeper2020reinforcement} and \citet{zakka2022xirl}. However, we make two changes, (1) Instead of using a fixed goal position, we use a randomized goal position and learn a goal-conditioned policy and (2) we do not use the sparse environment reward and instead only use the learned rewards for GraphIRL and learning-based baselines. Figure \ref{fig:mujoco_pusher} presents our results, we note that GraphIRL achieves slightly lower success rate than the task-specific environment reward (e.g. GraphIRL $0.455$ vs Environment Reward $0.6133$). Further, all vision-based baselines perform significantly lower than GraphIRL (e.g. GraphIRL $0.455$ vs XIRL $0.125$ and TCN $0.005$). For all learning-based methods, we use the data from \citet{schmeckpeper2020reinforcement} as training demonstrations similar to \emph{Push} experiments conducted in Section \ref{sec:experiments}}.

\input{figures/appendix-G/robot-setup}
\section{Robot Setup}
\label{sec:robot_setup}
We use a Ufactory xArm 7 robot for our real robot experiments. As shown in Figure \ref{fig:real-robot-setup}, we use a fixed third-person camera and an egocentric camera that is attached above the robot's gripper. Example images of the egocentric and third-person camera feeds passed to the RL agent are shown in Figure \ref{fig:real-robot-setup} (c) and Figure \ref{fig:real-robot-setup} (d).

\end{appendices}

\end{document}

%% file: tables/xmagical/overview-long.tex
\begin{figure}[t!]
    \centering
    \begin{minipage}{\textwidth}%
        \centering
        \textbf{\textit{Standard Environment}~~~~~~~~~~~~~~~~~~~~~~~~~~~~~~~~~~~~~~~~~~~\textit{Diverse Environment}}\vspace{0.2in}\\
        \begin{minipage}{0.11\textwidth}
            \centering
            \textbf{Gripper}\vspace{0.005in}\\
            \includegraphics[width=\textwidth]{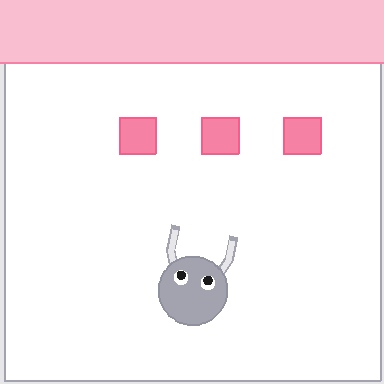}\vspace{0.005in}\\
        \end{minipage}\hspace{0.01in}
        \begin{minipage}{0.11\textwidth}
            \centering
            \textbf{S-stick}\vspace{0.03in}\\
            \includegraphics[width=\textwidth]{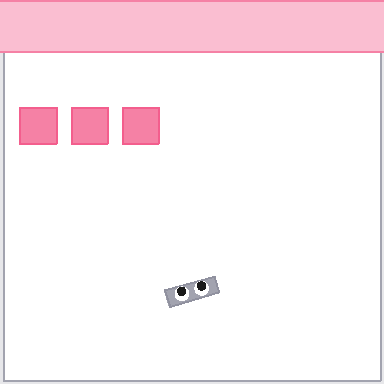}\vspace{0.005in}\\
        \end{minipage}\hspace{0.01in}
        \begin{minipage}{0.11\textwidth}
            \centering
            \textbf{M-stick}\vspace{0.03in}\\
            \includegraphics[width=\textwidth]{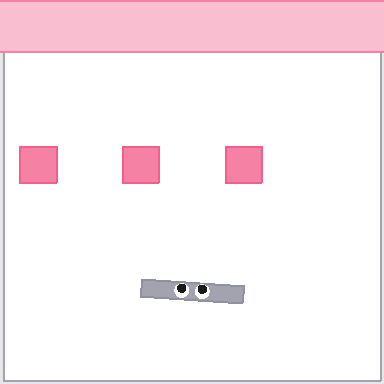}\vspace{0.005in}\\
        \end{minipage}\hspace{0.01in}
        \begin{minipage}{0.11\textwidth}
            \centering
            \textbf{L-stick}\vspace{0.02in}\\
            \includegraphics[width=\textwidth]{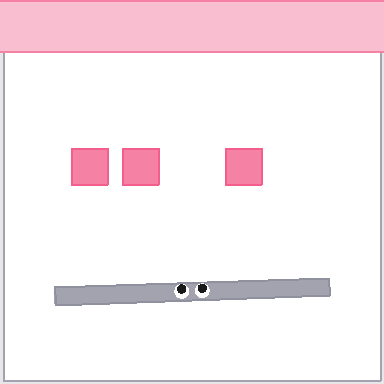}\vspace{0.005in}\\
        \end{minipage} \hspace{0.2in}
        \begin{minipage}{0.11\textwidth}
            \centering
            \textbf{Gripper}\vspace{0.005in}\\
            \includegraphics[width=\textwidth]{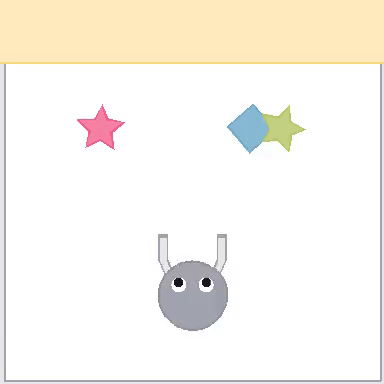}\vspace{0.005in}\\
        \end{minipage}\hspace{0.01in}
        \begin{minipage}{0.11\textwidth}
            \centering
            \textbf{S-stick}\vspace{0.03in}\\
            \includegraphics[width=\textwidth]{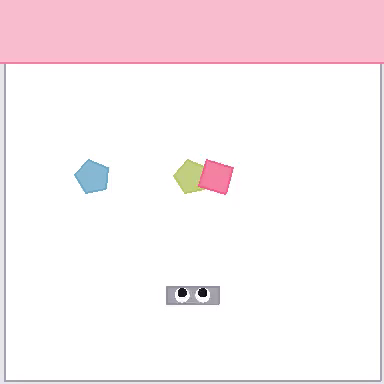}\vspace{0.005in}\\
        \end{minipage}\hspace{0.01in}
        \begin{minipage}{0.11\textwidth}
            \centering
            \textbf{M-stick}\vspace{0.03in}\\
            \includegraphics[width=\textwidth]{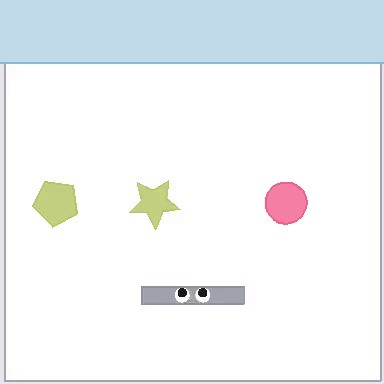}\vspace{0.005in}\\
        \end{minipage}\hspace{0.01in}
        \begin{minipage}{0.11\textwidth}
            \centering
            \textbf{L-stick}\vspace{0.02in}\\
            \includegraphics[width=\textwidth]{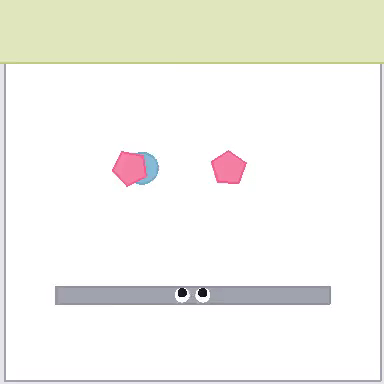}\vspace{0.005in}\\
        \end{minipage}
    \end{minipage}
    \caption{\textbf{Overview of X-MAGICAL task variants.} We consider two environment variants and four embodiments for our simulated sweeping task experiments. Our work assesses the performance of IRL algorithms in both the \textit{Diverse} and \textit{Standard} environments across all four embodiments in the \textit{Same-Embodiment} and \textit{Cross-Embodiment settings}.}
    \label{fig:xmagical-variants}
     \vspace{-0.05in}
\end{figure}

%% file: tables/robot-real.tex
\begin{table*}[t!]
\begin{minipage}[c]{1.0 \textwidth}
\centering
\resizebox{0.525\textwidth}{!}{%
\begin{tabular}{lccc}
\specialrule{1pt}{1pt}{1pt}
\textit{Real}  &  {XIRL} & {Env. Reward} & {GraphIRL (Ours)}\\
\midrule
Push	& $0.27$ & ${0.47}$ & $\mathbf{0.60}$ \\
Reach & $0.26$  & $\mathbf{0.93}$ & ${0.86}$  \\
Peg in Box &  $0.06$ & $\mathbf{0.60}$ & ${0.53}$ \\
\specialrule{1pt}{1pt}{1pt}
\end{tabular}
}
\caption{\textbf{Real robot experiments.} Success rate on robot manipulation tasks on physical hardware. We evaluate each method for 15 trials using a fixed set of goal and start state configurations. Best results are in \textbf{bold}.}
\label{tab:robot_real}
\end{minipage}\hfill
\vspace{-0.2in}
\end{table*}

%% file: tables/ablation.tex
\begin{table}[t]
\begin{minipage}[c]{0.48\linewidth}
\centering
\begin{tabular}{l|cc}

\specialrule{1pt}{1pt}{1pt}

Variant  &  {Success Rate}  \\
\midrule
  MLP & $0.61 \scriptstyle{\pm 0.116}$	   \\
 IN &  $\mathbf{0.804 \scriptstyle{\pm 0.054}}$   \\

\specialrule{1pt}{1pt}{1pt}
\end{tabular}
\vspace{0.05in}
\caption{Impact of modelling
object-object interaction on \emph{Push} task. \textbf{MLP};
Multi-layer perceptron and \textbf{IN}: Spatial Interaction
Network Encoder. Results averaged over 5
seeds. Best results are in \textbf{bold}.}
\label{tab:ablation_interactions}

\end{minipage}\hfill
\begin{minipage}[c]{0.48\linewidth}
\centering
\begin{tabular}{l|c}

\specialrule{1pt}{1pt}{1pt}

\% Videos Used  &  {Success Rate}  \\
\midrule
25\% &  $0.670 \scriptstyle{\pm 0.176}$  \\
50\% &  $0.755 \scriptstyle{\pm 0.019}$ \\
75\% &  $0.776 \scriptstyle{\pm 0.04}$  \\
100\% & $\mathbf{0.804 \scriptstyle{\pm 0.054}}$\\

\specialrule{1pt}{1pt}{1pt}
\end{tabular}
\vspace{0.05in}
\caption{Impact of reducing number of pretraining demonstrations on \textit{Push} task. Results averaged over 5 seeds. Best results are in \textbf{bold}.}
\label{tab:ablation_data}

\end{minipage}
\vspace{-0.2in}
\end{table}

%% file: figures/appendix-A/pegbox/2/pegbox_success.tex
\begin{figure}[ht!]
    \centering
    \begin{minipage}{0.49\textwidth}%
        \centering
        \begin{minipage}{0.025\textwidth}
            \centering
        \end{minipage}\hspace{0.01in}
        \begin{minipage}{0.93\textwidth}
            \centering
            \includegraphics[width=0.9\textwidth]{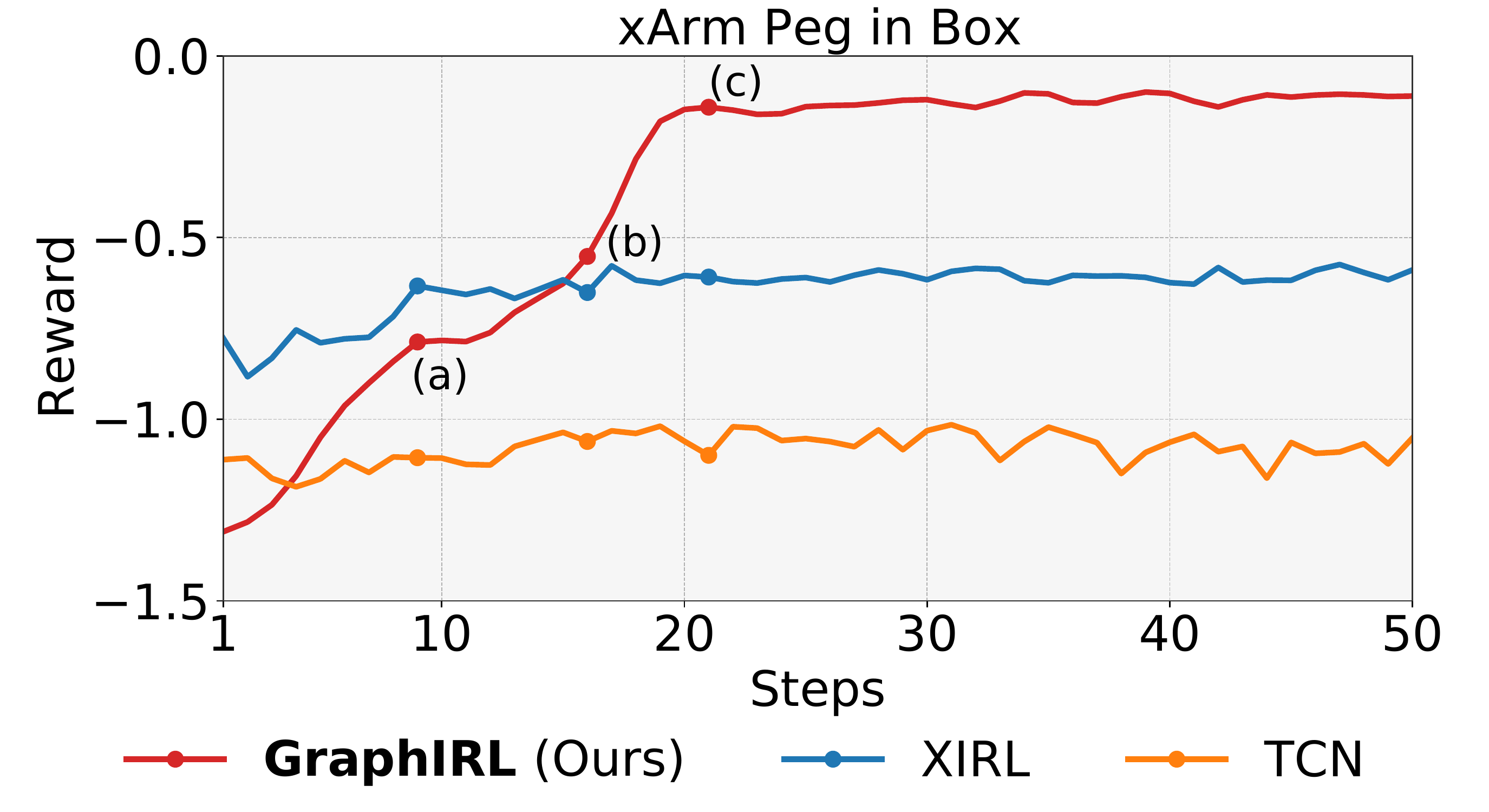}\vspace{0.005in}\\
        \end{minipage}\vspace{0.05in}
        \begin{minipage}{0.3\textwidth}
            \centering
            \includegraphics[width=\textwidth]{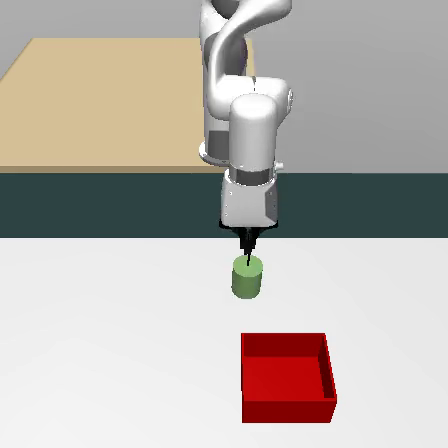}\vspace{0.005in}\hspace{0.05in}\\
            \textbf{(a)}\vspace{0.03in}\\
        \end{minipage}\hspace{0.05in}
        \begin{minipage}{0.3\textwidth}
            \centering
            \includegraphics[width=\textwidth]{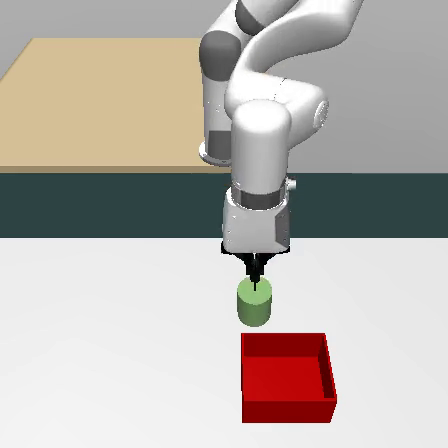}\vspace{0.005in}\hspace{0.05in}\\
            \textbf{(b)}\vspace{0.03in}\\
        \end{minipage}\hspace{0.05in}
        \begin{minipage}{0.3\textwidth}
            \centering
            \includegraphics[width=\textwidth]{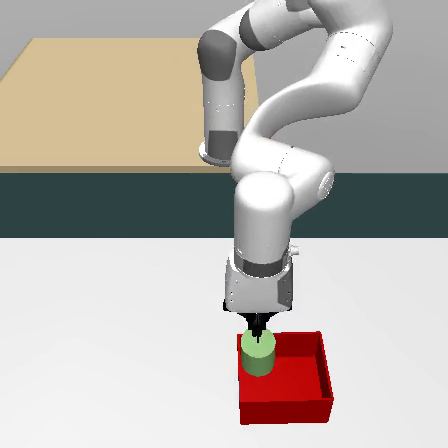}\vspace{0.005in}\hspace{0.05in}\\
            \textbf{(c)}\vspace{0.03in}\\
        \end{minipage}\hspace{0.05in}
    \end{minipage}\hspace{0.05in}%
    \begin{minipage}{0.49\textwidth}%
        \centering
        \begin{minipage}{0.025\textwidth}
            \centering
        \end{minipage}\hspace{0.01in}
        \begin{minipage}{0.93\textwidth}
            \centering
            \includegraphics[width=0.9\textwidth]{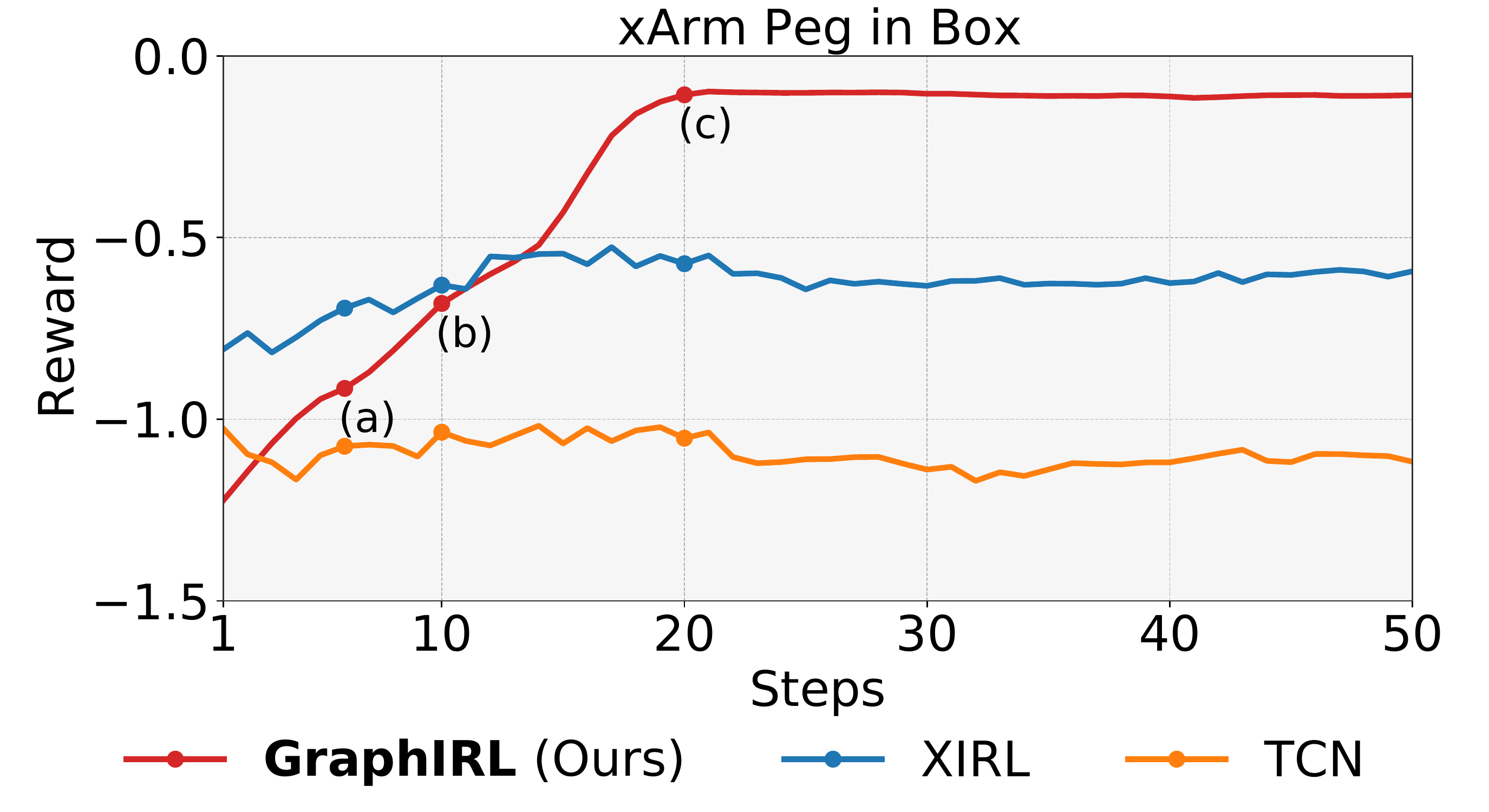}\vspace{0.005in}\\
        \end{minipage}\vspace{0.05in}
        \begin{minipage}{0.3\textwidth}
            \centering
            \includegraphics[width=\textwidth]{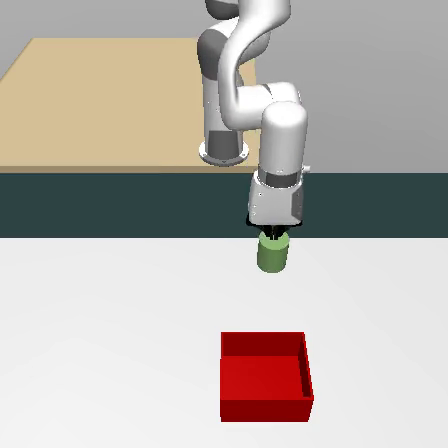}\vspace{0.005in}\hspace{0.05in}\\
            \textbf{(a)}\vspace{0.03in}\\
        \end{minipage}\hspace{0.05in}
        \begin{minipage}{0.3\textwidth}
            \centering
            \includegraphics[width=\textwidth]{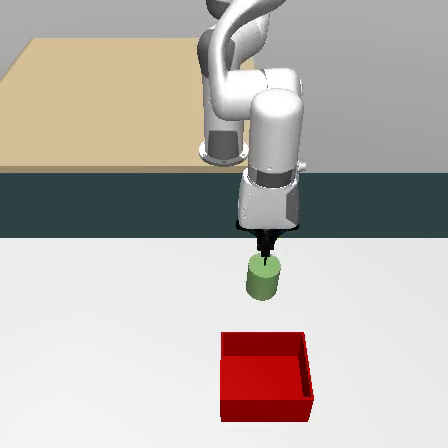}\vspace{0.005in}\hspace{0.05in}\\
            \textbf{(b)}\vspace{0.03in}\\
        \end{minipage}\hspace{0.05in}
        \begin{minipage}{0.3\textwidth}
            \centering
            \includegraphics[width=\textwidth]{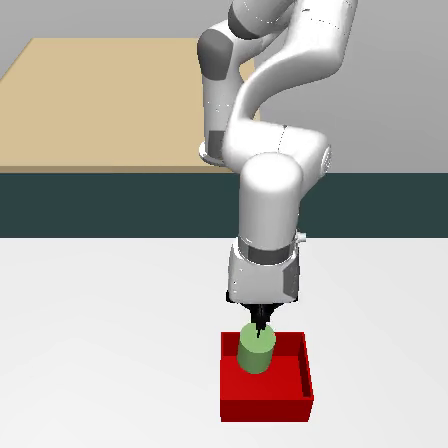}\vspace{0.005in}\hspace{0.05in}\\
            \textbf{(c)}\vspace{0.03in}\\
        \end{minipage}\hspace{0.05in}
    \end{minipage}\hspace{0.05in}
    \caption{\textbf{Peg in Box Task Progress: Success}. For the \textit{Peg in Box} task setting, we find that \textit{GraphIRL} provides an accurate measurement of task progress. Pictured are video frames \textbf{(a)}, \textbf{(b)}, \textbf{(c)} which denote critical points of task progress. Task progress is measured using video frames from a 50-step evaluation episode.}
    \label{fig:pegbox-success}
\end{figure}

%% file: figures/appendix-A/pegbox/0/pegbox_failure.tex
\begin{figure}[ht!]
    \centering
    \begin{minipage}{0.5\textwidth}%
        \centering
        \begin{minipage}{\textwidth}
            \centering
            \includegraphics[width=0.9\textwidth]{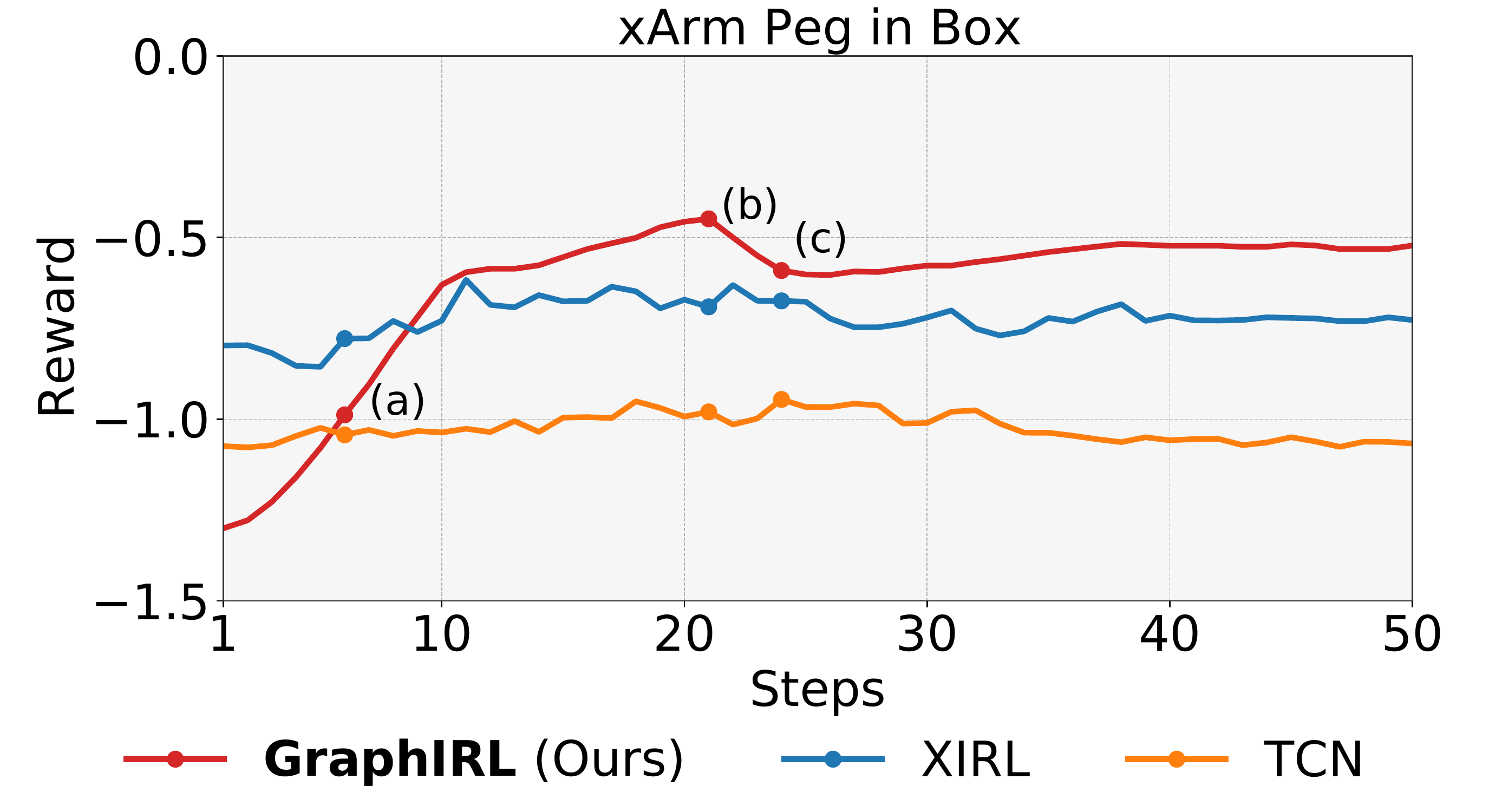}\vspace{0.005in}\\
        \end{minipage}\vspace{0.05in}
        \begin{minipage}{0.3\textwidth}
            \centering
            \includegraphics[width=\textwidth]{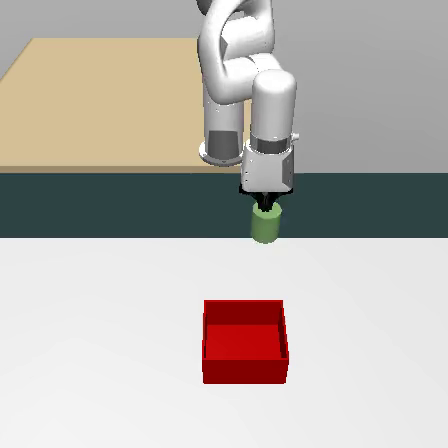}\vspace{0.005in}\hspace{0.05in}\\
            \textbf{(a)}\vspace{0.03in}\\
        \end{minipage}\hspace{0.05in}
        \begin{minipage}{0.3\textwidth}
            \centering
            \includegraphics[width=\textwidth]{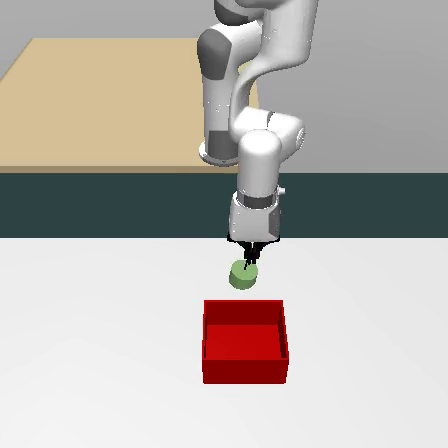}\vspace{0.005in}\hspace{0.05in}\\
            \textbf{(b)}\vspace{0.03in}\\
        \end{minipage}\hspace{0.05in}
        \begin{minipage}{0.3\textwidth}
            \centering
            \includegraphics[width=\textwidth]{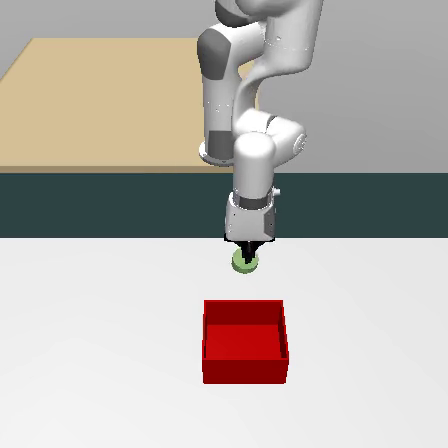}\vspace{0.005in}\hspace{0.05in}\\
            \textbf{(c)}\vspace{0.03in}\\
        \end{minipage}\hspace{0.05in}
    \end{minipage}\hspace{0.05in}
    \caption{\textbf{Peg in Box Task Progress: Failure}. \textit{GraphIRL} measures positive task progress until the peg goes into the table, a critical failure point for the task. The physical interaction between the peg and table is unnatural, and our method succeeds in recognizing this.}
    \label{fig:pegbox-failure}
\end{figure}

%% file: figures/appendix-A/push/2/push_success.tex
\begin{figure}[ht!]
    \centering
    \begin{minipage}{0.49\textwidth}%
        \centering
        \begin{minipage}{0.025\textwidth}
            \centering
        \end{minipage}\hspace{0.01in}
        \begin{minipage}{0.93\textwidth}
            \centering
            \includegraphics[width=0.9\textwidth]{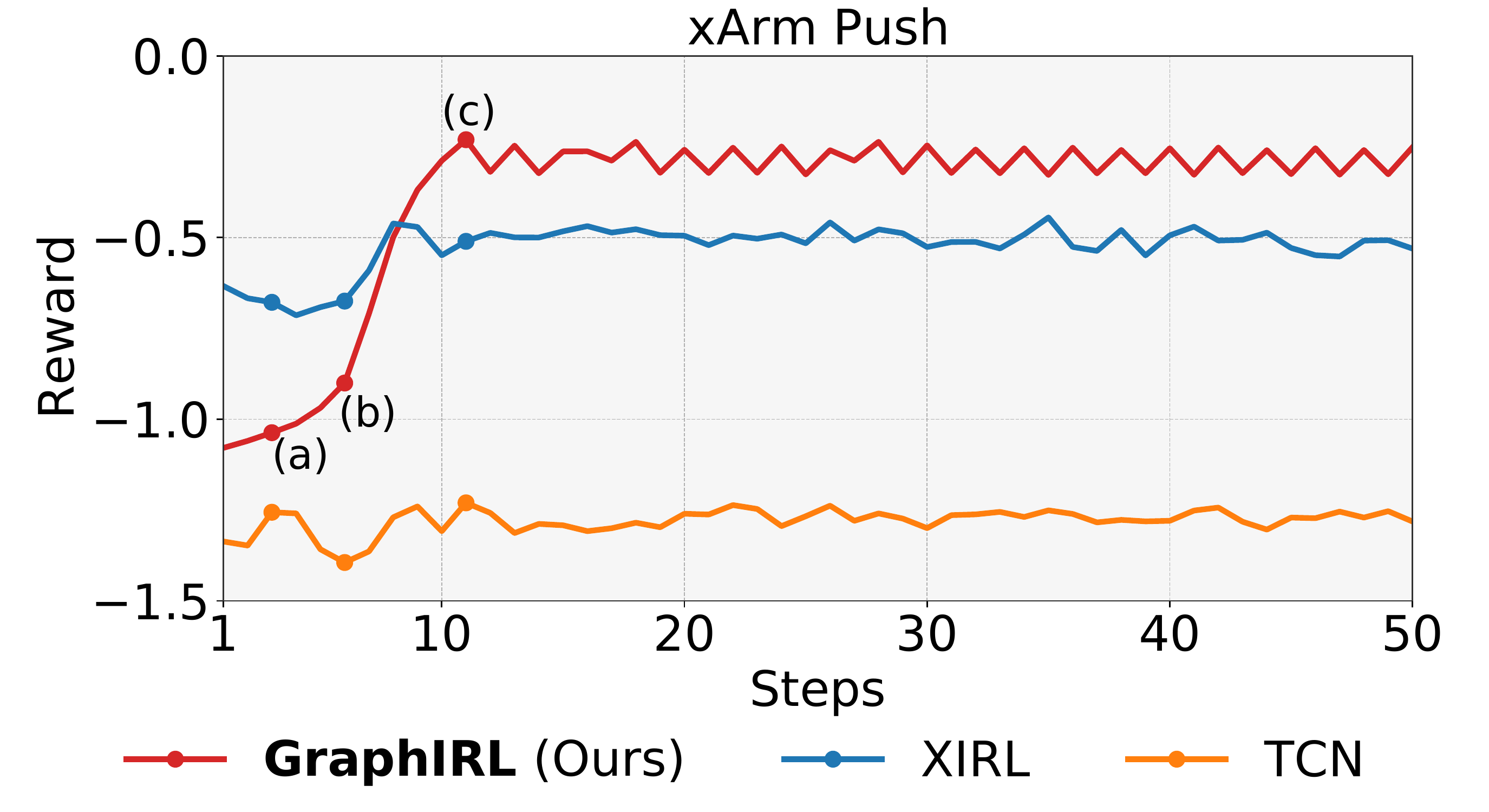}\vspace{0.005in}\\
        \end{minipage}\vspace{0.05in}
        \begin{minipage}{0.3\textwidth}
            \centering
            \includegraphics[width=\textwidth]{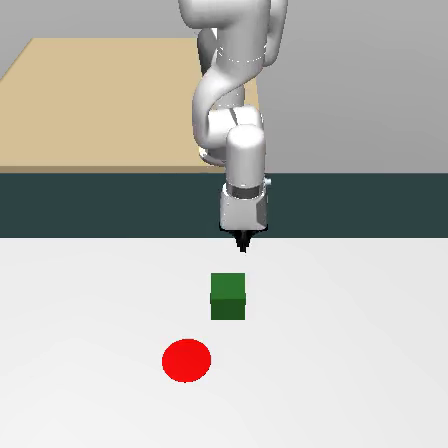}\vspace{0.005in}\hspace{0.05in}\\
            \textbf{(a)}\vspace{0.03in}\\
        \end{minipage}\hspace{0.05in}
        \begin{minipage}{0.3\textwidth}
            \centering
            \includegraphics[width=\textwidth]{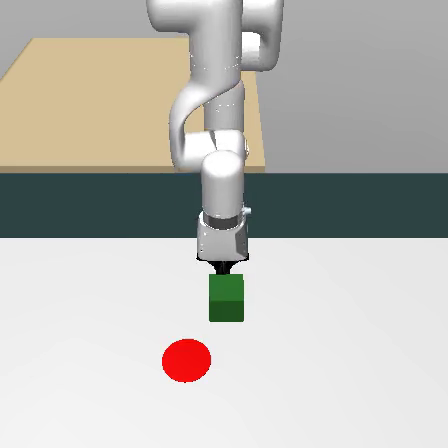}\vspace{0.005in}\hspace{0.05in}\\
            \textbf{(b)}\vspace{0.03in}\\
        \end{minipage}\hspace{0.05in}
        \begin{minipage}{0.3\textwidth}
            \centering
            \includegraphics[width=\textwidth]{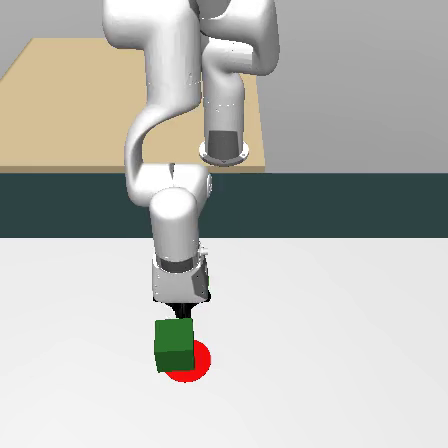}\vspace{0.005in}\hspace{0.05in}\\
            \textbf{(c)}\vspace{0.03in}\\
        \end{minipage}\hspace{0.05in}
    \end{minipage}\hspace{0.05in}%
    \begin{minipage}{0.49\textwidth}%
        \centering
        \begin{minipage}{0.025\textwidth}
            \centering
        \end{minipage}\hspace{0.01in}
        \begin{minipage}{0.93\textwidth}
            \centering
            \includegraphics[width=0.9\textwidth]{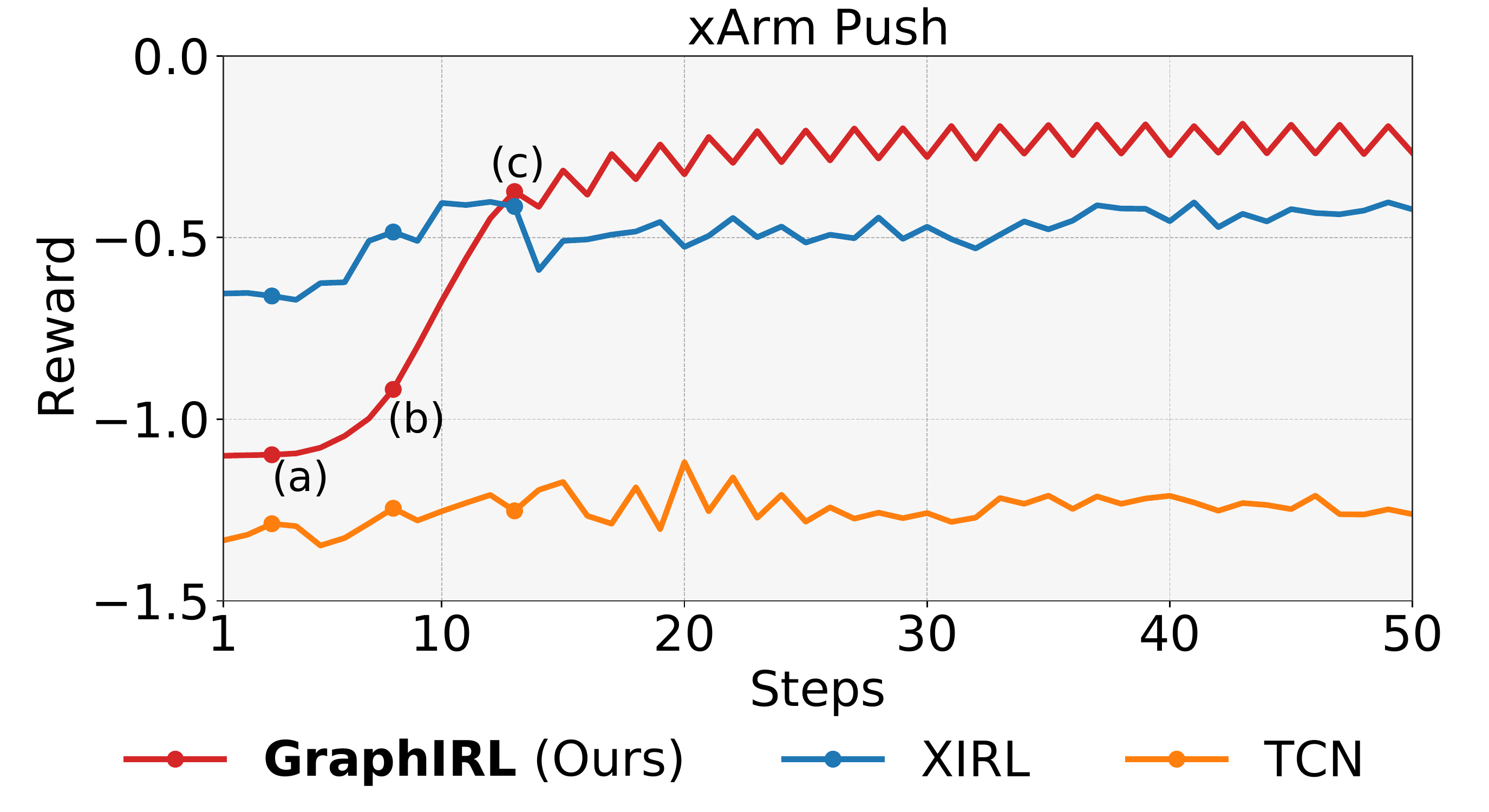}\vspace{0.005in}\\
        \end{minipage}\vspace{0.05in}
        \begin{minipage}{0.3\textwidth}
            \centering
            \includegraphics[width=\textwidth]{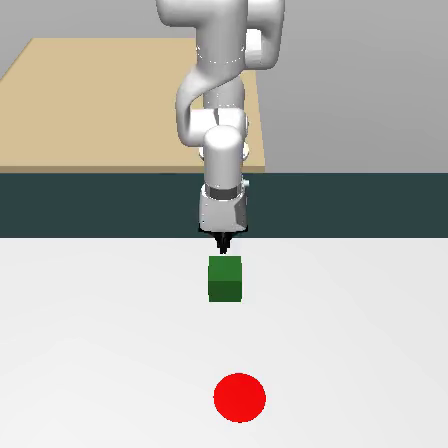}\vspace{0.005in}\hspace{0.05in}\\
            \textbf{(a)}\vspace{0.03in}\\
        \end{minipage}\hspace{0.05in}
        \begin{minipage}{0.3\textwidth}
            \centering
            \includegraphics[width=\textwidth]{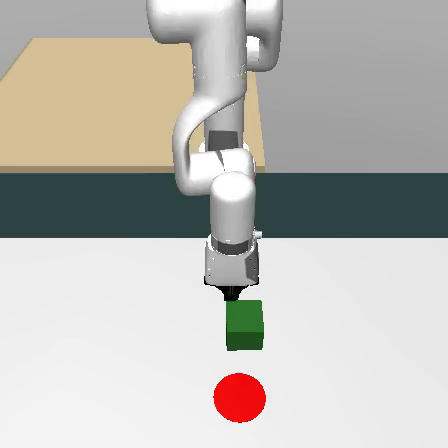}\vspace{0.005in}\hspace{0.05in}\\
            \textbf{(b)}\vspace{0.03in}\\
        \end{minipage}\hspace{0.05in}
        \begin{minipage}{0.3\textwidth}
            \centering
            \includegraphics[width=\textwidth]{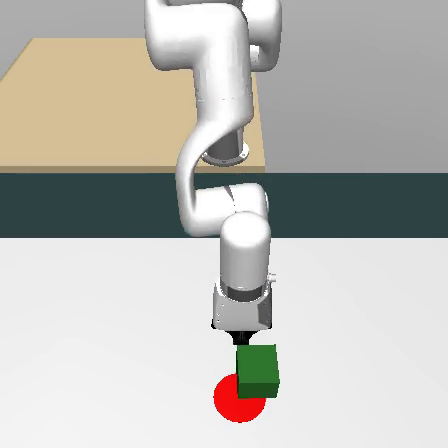}\vspace{0.005in}\hspace{0.05in}\\
            \textbf{(c)}\vspace{0.03in}\\
        \end{minipage}\hspace{0.05in}
    \end{minipage}\hspace{0.05in}
    \caption{\textbf{Push Task Progress: Success}. The \textit{Push} task setting is often completed within the first 10 steps of the evaluation episode, and as shown between Steps 1 through 10 in both success examples, \textit{GraphIRL} measures high task progress. XIRL and TCN on the other hand, incorrectly show much lower task progress.}
    \label{fig:push-success}
\end{figure}

%% file: figures/appendix-A/push/0/push_failure.tex
\begin{figure}[ht!]
    \centering
    \begin{minipage}{0.5\textwidth}%
        \centering
        \begin{minipage}{0.025\textwidth}
            \centering
        \end{minipage}\hspace{0.01in}
        \begin{minipage}{0.93\textwidth}
            \centering
            \includegraphics[width=0.9\textwidth]{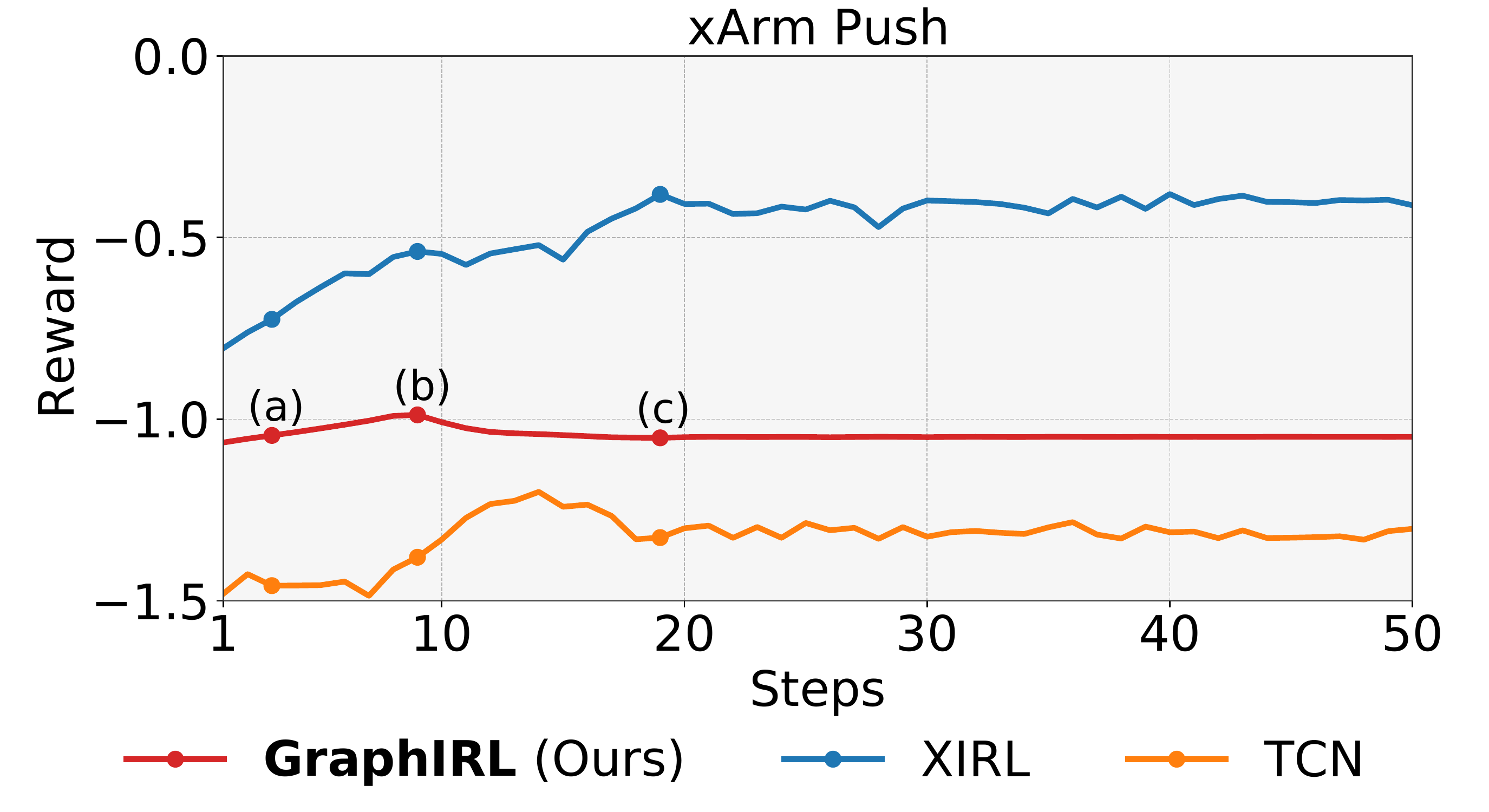}\vspace{0.005in}\\
        \end{minipage}\vspace{0.05in}
        \begin{minipage}{0.3\textwidth}
            \centering
            \includegraphics[width=\textwidth]{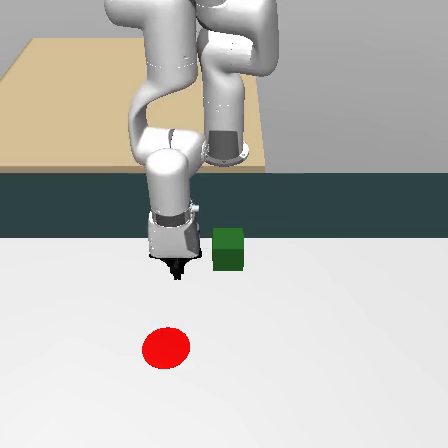}\vspace{0.005in}\hspace{0.05in}\\
            \textbf{(a)}\vspace{0.03in}\\
        \end{minipage}\hspace{0.05in}
        \begin{minipage}{0.3\textwidth}
            \centering
            \includegraphics[width=\textwidth]{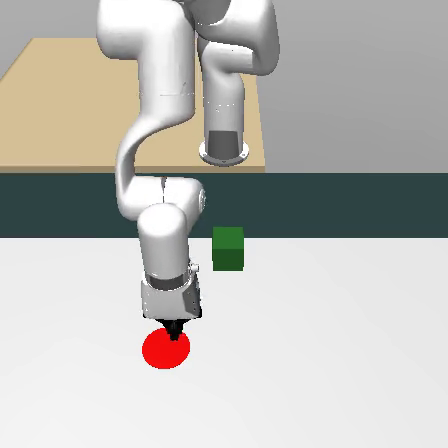}\vspace{0.005in}\hspace{0.05in}\\
            \textbf{(b)}\vspace{0.03in}\\
        \end{minipage}\hspace{0.05in}
        \begin{minipage}{0.3\textwidth}
            \centering
            \includegraphics[width=\textwidth]{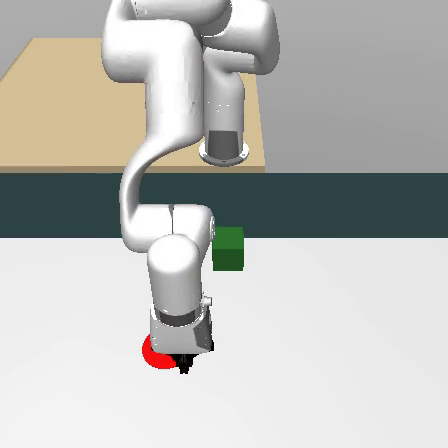}\vspace{0.005in}\hspace{0.05in}\\
            \textbf{(c)}\vspace{0.03in}\\
        \end{minipage}\hspace{0.05in}
    \end{minipage}\hspace{0.05in}
    \caption{\textbf{Push Task Progress: Failure}. \textit{GraphIRL}'s understanding of object relationships is made clear in this \textit{Push} task failure, since without any forward movement of the box object towards the goal, no positive task progress is made. Other baselines rely on direct visual input of the task, and because of this, they inaccurately align visual states \textbf{(a)}, \textbf{(b)}, \textbf{(c)} of the task with positive task progress.}
    \label{fig:push-failure}
\end{figure}

%% file: figures/appendix-A/reach/2/reach_success.tex
\begin{figure}[ht!]
    \centering
    \begin{minipage}{0.49\textwidth}%
        \centering
        \begin{minipage}{0.025\textwidth}
            \centering
        \end{minipage}\hspace{0.01in}
        \begin{minipage}{0.93\textwidth}
            \centering
            \includegraphics[width=0.9\textwidth]{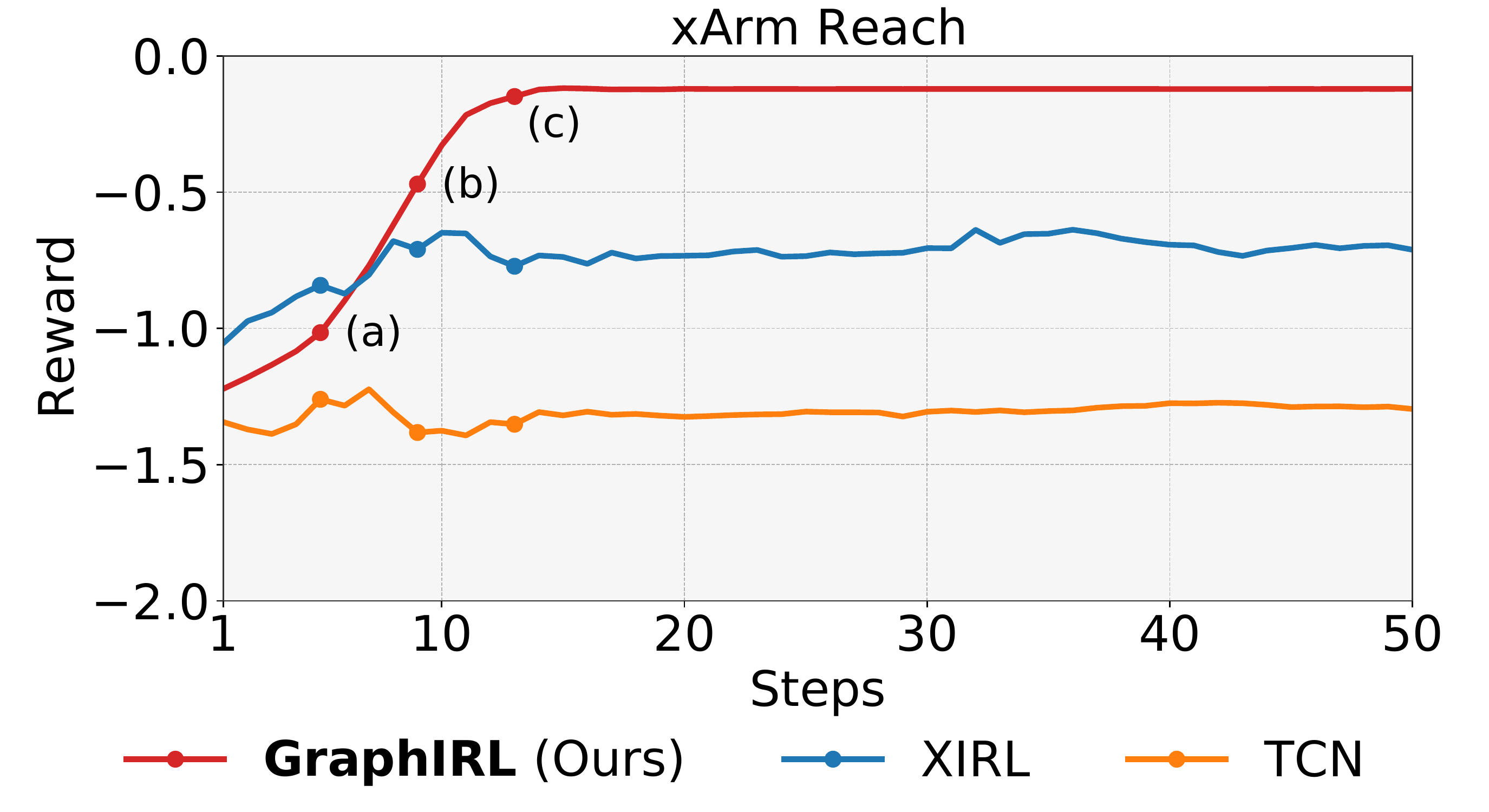}\vspace{0.005in}\\
        \end{minipage}\vspace{0.05in}
        \begin{minipage}{0.3\textwidth}
            \centering
            \includegraphics[width=\textwidth]{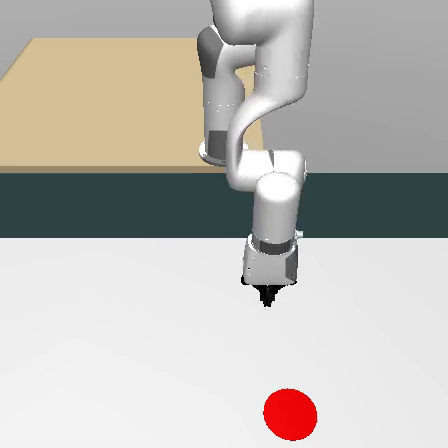}\vspace{0.005in}\hspace{0.05in}\\
            \textbf{(a)}\vspace{0.03in}\\
        \end{minipage}\hspace{0.05in}
        \begin{minipage}{0.3\textwidth}
            \centering
            \includegraphics[width=\textwidth]{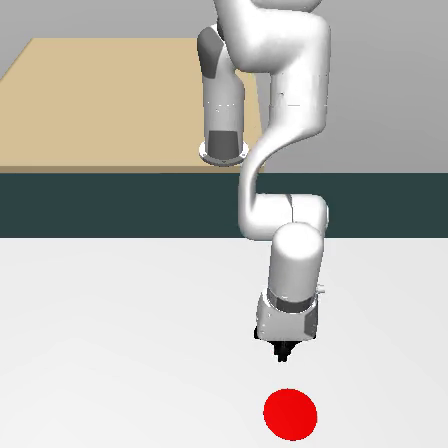}\vspace{0.005in}\hspace{0.05in}\\
            \textbf{(b)}\vspace{0.03in}\\
        \end{minipage}\hspace{0.05in}
        \begin{minipage}{0.3\textwidth}
            \centering
            \includegraphics[width=\textwidth]{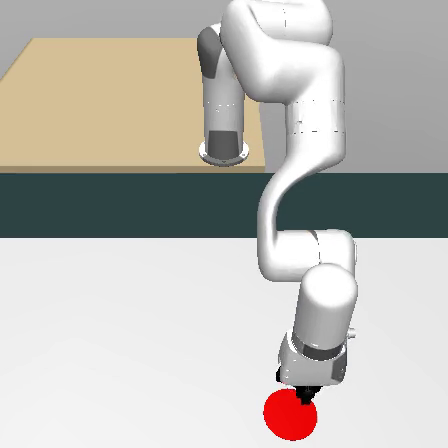}\vspace{0.005in}\hspace{0.05in}\\
            \textbf{(c)}\vspace{0.03in}\\
        \end{minipage}\hspace{0.05in}
    \end{minipage}\hspace{0.05in}%
    \begin{minipage}{0.49\textwidth}%
        \centering
        \begin{minipage}{0.025\textwidth}
            \centering
        \end{minipage}\hspace{0.01in}
        \begin{minipage}{0.93\textwidth}
            \centering
            \includegraphics[width=0.9\textwidth]{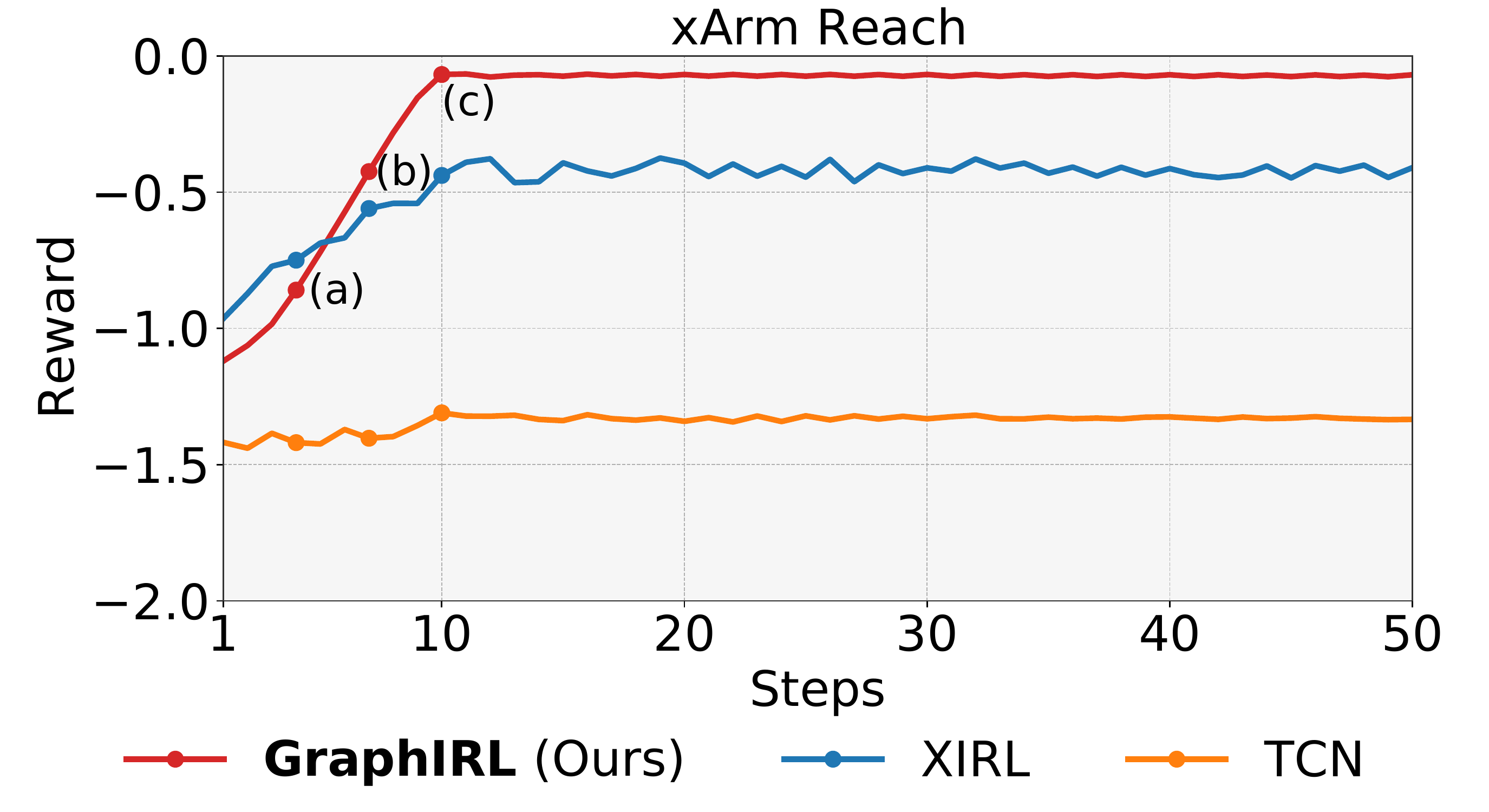}\vspace{0.005in}\\
        \end{minipage}\vspace{0.05in}
        \begin{minipage}{0.3\textwidth}
            \centering
            \includegraphics[width=\textwidth]{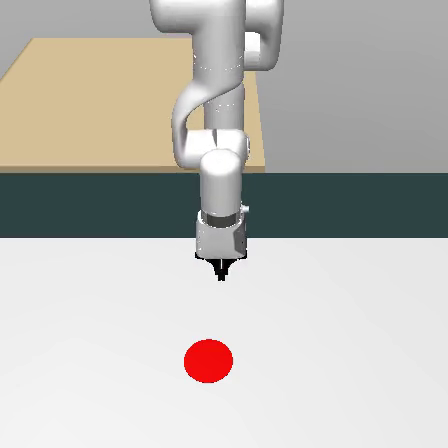}\vspace{0.005in}\hspace{0.05in}\\
            \textbf{(a)}\vspace{0.03in}\\
        \end{minipage}\hspace{0.05in}
        \begin{minipage}{0.3\textwidth}
            \centering
            \includegraphics[width=\textwidth]{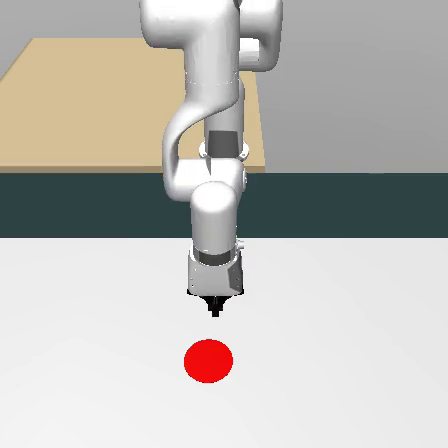}\vspace{0.005in}\hspace{0.05in}\\
            \textbf{(b)}\vspace{0.03in}\\
        \end{minipage}\hspace{0.05in}
        \begin{minipage}{0.3\textwidth}
            \centering
            \includegraphics[width=\textwidth]{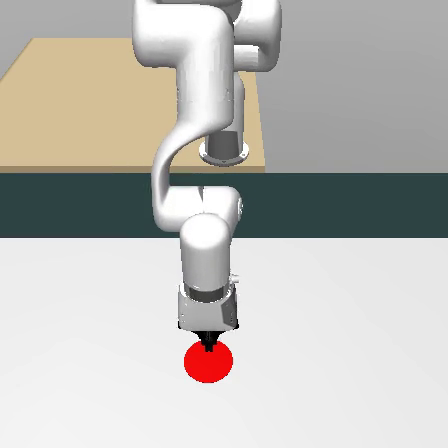}\vspace{0.005in}\hspace{0.05in}\\
            \textbf{(c)}\vspace{0.03in}\\
        \end{minipage}\hspace{0.05in}
    \end{minipage}\hspace{0.05in}
    \caption{\textbf{Reach Task Progress: Success}. In the \textit{Reach} task setting, positive task progress is measured by \textit{GraphIRL} with forward movement of the end-effector gripper towards the goal location. The image frames \textbf{(a)}, \textbf{(b)}, \textbf{(c)} reflect the alignment between measured task progress and visual state of the task.}
    \label{fig:reach-success}
\end{figure}

%% file: figures/appendix-A/reach/0/reach_failure.tex
\begin{figure}[ht!]
    \centering
    \begin{minipage}{0.5\textwidth}%
        \centering
        \begin{minipage}{0.025\textwidth}
            \centering
        \end{minipage}\hspace{0.01in}
        \begin{minipage}{0.93\textwidth}
            \centering
            \includegraphics[width=0.9\textwidth]{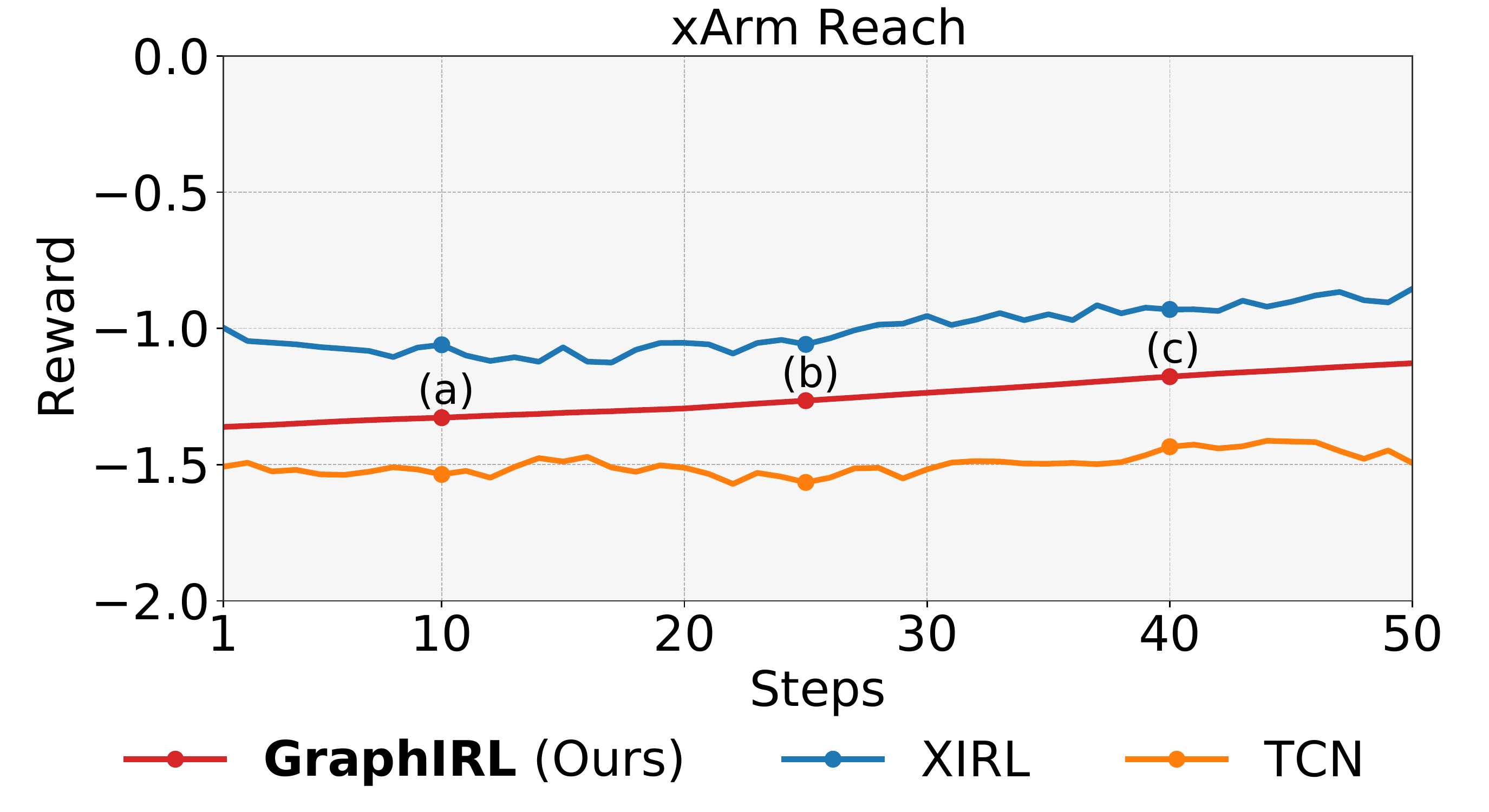}\vspace{0.005in}\\
        \end{minipage}\vspace{0.05in}
        \begin{minipage}{0.3\textwidth}
            \centering
            \includegraphics[width=\textwidth]{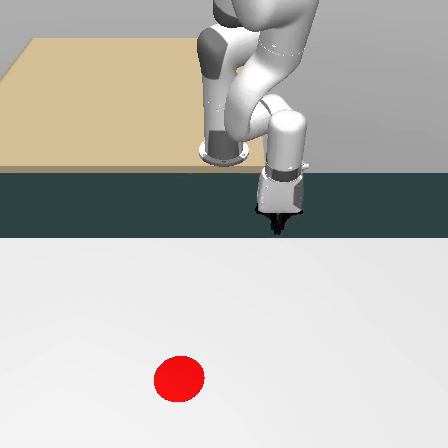}\vspace{0.005in}\hspace{0.05in}\\
            \textbf{(a)}\vspace{0.03in}\\
        \end{minipage}\hspace{0.05in}
        \begin{minipage}{0.3\textwidth}
            \centering
            \includegraphics[width=\textwidth]{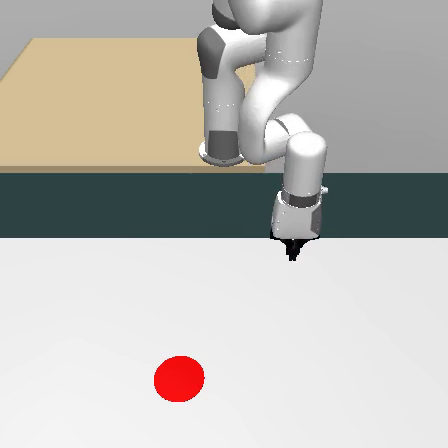}\vspace{0.005in}\hspace{0.05in}\\
            \textbf{(b)}\vspace{0.03in}\\
        \end{minipage}\hspace{0.05in}
        \begin{minipage}{0.3\textwidth}
            \centering
            \includegraphics[width=\textwidth]{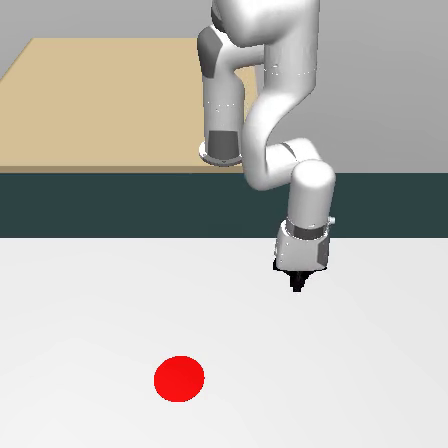}\vspace{0.005in}\hspace{0.05in}\\
            \textbf{(c)}\vspace{0.03in}\\
        \end{minipage}\hspace{0.05in}
    \end{minipage}\hspace{0.05in}
    \caption{\textbf{Reach Task Progress: Failure}. Our \textit{GraphIRL} method measures an approximately linear task progress in this failure example for \textit{Reach}. The gripper's distance to the goal region is indeed minimized over time, though since it does not get within close-enough distance to the goal, the measured task progress is lower compared to success examples shown in Figure \ref{fig:reach-success}.}
    \label{fig:reach-failure}
\end{figure}

%% file: tables/hyperparameters.tex
\begin{table}[ht]
\begin{minipage}[c]{1.0\linewidth}
\centering
\begin{tabular}{l|cc}

\specialrule{1pt}{1pt}{1pt}

Hyperparameter  &  Value  \\
\midrule
  \# of sampled frames & $90$	   \\
 Batch Size &  $2$   \\
 Learning Rate & $10^{-5}$ \\
 Weight Decay & $10^{-5}$ \\
 \# of training iterations & $12000$ \\ 
 Embedding Size & $128$ \\
 Softmax Temperature & $0.1$ \\

\specialrule{1pt}{1pt}{1pt}
\end{tabular}
\vspace{0.05in}
\caption{Hyperparameters for Representation Learning with GraphIRL.}
\label{tab:hyperparameters}

\end{minipage}\hfill
\end{table}

%% file: figures/appendix-G/robot-setup.tex
\begin{figure}[!ht]
    \centering
    \begin{minipage}{0.8\textwidth}%
        \centering
        \begin{minipage}{0.3\textwidth}
            \centering
            \includegraphics[width=\textwidth]{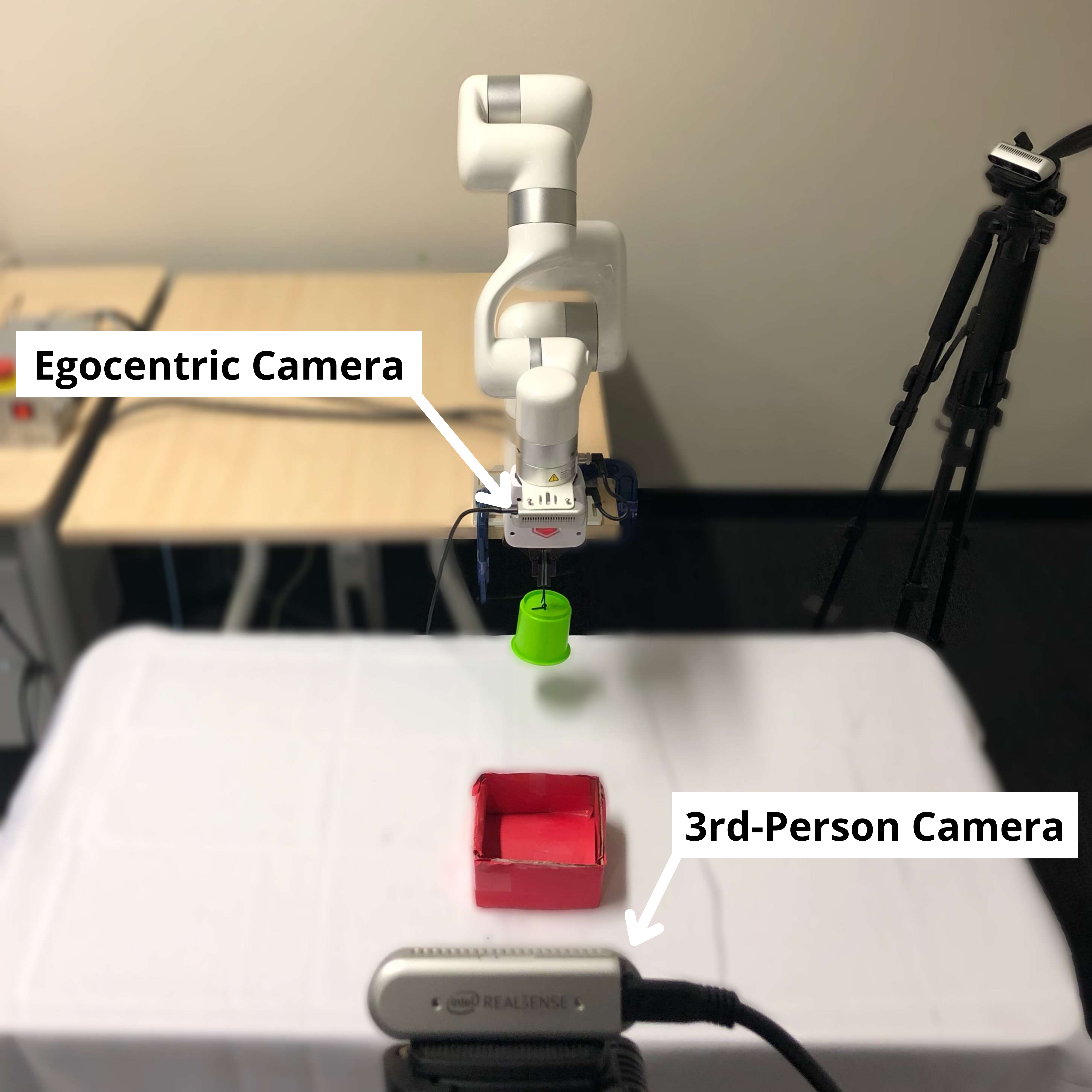}\vspace{0.005in}\hspace{0.05in}\\
            \textbf{(a)}\vspace{0.03in}\\
        \end{minipage}\hspace{0.05in}
        \begin{minipage}{0.3\textwidth}
            \centering
            \includegraphics[width=\textwidth]{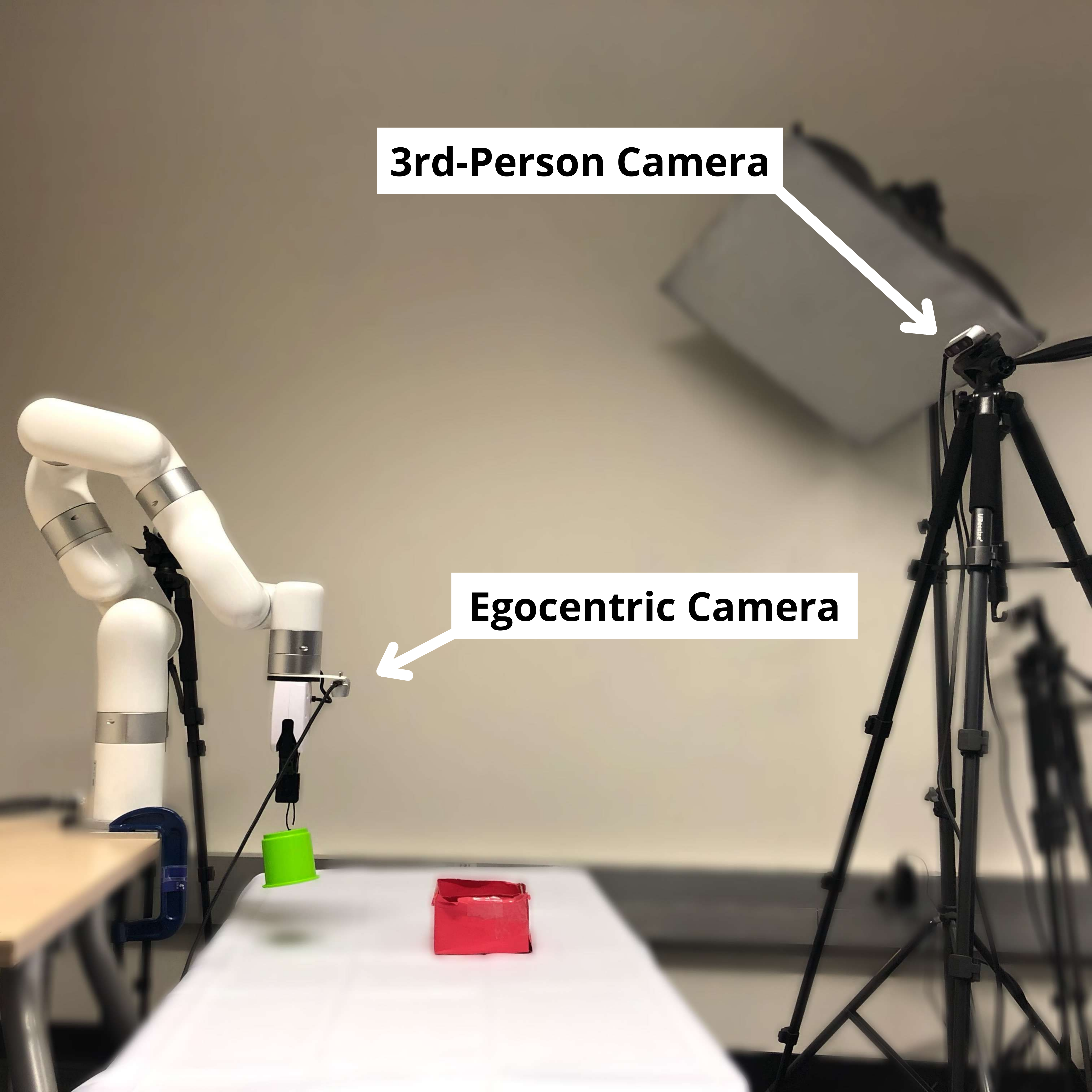}\vspace{0.005in}\hspace{0.05in}\\
            \textbf{(b)}\vspace{0.03in}\\
        \end{minipage}\hspace{0.05in}
        
        \begin{minipage}{0.3\textwidth}
            \centering
            \includegraphics[width=\textwidth]{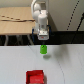}\vspace{0.005in}\hspace{0.05in}\\
            \textbf{(c)}\vspace{0.03in}\\
        \end{minipage}\hspace{0.05in}
        \begin{minipage}{0.3\textwidth}
            \centering
            \includegraphics[width=\textwidth]{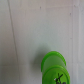}\vspace{0.005in}\hspace{0.05in}\\
            \textbf{(d)}\vspace{0.03in}\\
        \end{minipage}\hspace{0.05in}
    \end{minipage}\hspace{0.05in}
    \caption{\textbf{Real Robot Setup}. In \textbf{(a)} and \textbf{(b)}, we provide images of our real-world environment for the \textit{Peg in Box} task. We use a static third-person camera and an egocentric camera which moves with the arm while completing the task. Pictured in \textbf{(c)} and \textbf{(d)} are single image frames captured by our third-person and egocentric cameras.}
    \label{fig:real-robot-setup}
\end{figure}